\documentclass[acmlarge]{acmart}

\makeatletter
\newcommand{\myconfshort}{\acmConference@shortname}
\newcommand{\myconffull}{\acmConference@name}
\newcommand{\myconfdate}{\acmConference@date}
\newcommand{\myconfloc}{\acmConference@venue}
\AtBeginDocument{
  \fancypagestyle{firstpagestyle}{
    \fancyhead{}%
    \fancyfoot[C]{}%
  }
  \fancyhf{}
  \fancyhead[LO]{\@headfootfont\shorttitle}%
  \fancyhead[RE]{\@headfootfont\@shortauthors}%
  \fancyhead[LE]{\@headfootfont\footnotesize \myconfshort, \myconfdate, \myconfloc}%
  \fancyhead[RO]{\@headfootfont\footnotesize \myconfshort, \myconfdate, \myconfloc}%
  \fancyfoot[C]{}%
}
\makeatother
\acmBooktitle{\conffull\@ (\confshort), \confdate, \confloc}

\AtBeginDocument{%
  }

\usepackage{graphicx}%
\usepackage{multirow}%
\usepackage{amsmath,amsfonts}%
\usepackage{amsthm}%
\usepackage{mathrsfs}%
\usepackage[title]{appendix}%
\usepackage{xcolor}%
\usepackage{textcomp}%
\usepackage{manyfoot}%
\usepackage{booktabs}%
\usepackage{listings}%
\usepackage{algorithmic}
\usepackage{graphicx}
\usepackage{textcomp}
\usepackage{xcolor}
\usepackage{etoolbox}
\usepackage{subcaption}
\usepackage{mathtools} %
\usepackage{booktabs} %
\usepackage{tikz} %
\usepackage{cleveref}

\usepackage{comment}
\usepackage{bm}
\usepackage{dsfont}
\usepackage{hyperref}
\usepackage{xfrac}

\usepackage{ulem}
\usepackage{etoolbox} %
\newtoggle{comments} %
\toggletrue{comments} %

\setcopyright{acmlicensed}

\copyrightyear{2026}
\acmYear{2026}
\setcopyright{cc}
\setcctype{by}
\acmConference[FAccT '26]{The 2026 ACM Conference on Fairness, Accountability, and Transparency}{June 25--28, 2026}{Montreal, QC, Canada}
\acmBooktitle{The 2026 ACM Conference on Fairness, Accountability, and Transparency (FAccT '26), June 25--28, 2026, Montreal, QC, Canada}
\acmDOI{10.1145/3805689.3812272}
\acmISBN{XXXXX}
\acmISBN{978-1-4503-XXXX-X/2018/06}

\DeclareMathOperator*{\argmax}{argmax}

\newcommand{\xhdr}[1]{\textbf{#1.}\;}

\begin{document}

\title{Fairness under uncertainty in sequential decisions}

\author{Michelle Seng Ah Lee}
\email{sal87@cam.ac.uk}
\orcid{0000-0001-7725-2503}
\affiliation{%
  \institution{University of Cambridge}
  \city{Cambridge}
  \country{UK}
}
\authornote{Authors Lee, Padh, and Watson contributed equally to this research}

\author{Kirtan Padh}
\orcid{}
\affiliation{%
  \institution{Technical University of Munich}
  \city{Munich}
  \country{Germany}}
\email{kirtan.padh@tum.de}
\authornotemark[1]

\author{David Watson}
\orcid{}
\affiliation{%
  \institution{King's College London}
  \city{London}
  \country{United Kingdom}
}
\email{david.watson@kcl.ac.uk}
\authornotemark[1]

\author{Niki Kilbertus}
\orcid{}
\affiliation{%
  \institution{Technical University of Munich, Helmholtz Munich, Munich Center for Machine Learning (MCML)}
  \city{Munich}
  \country{Germany}}
\email{niki.kilbertus@tum.de}

\author{Jatinder Singh}
\email{jat@compacctsys.net}
\orcid{0000-0002-5102-6564}
\affiliation{%
\department{{Research Centre Trust, UA Ruhr}}
\institution{University of Duisburg-Essen}
\city{Duisburg}
\country{Germany}
}
\affiliation{%
\institution{University of Cambridge}
\city{Cambridge}
\country{United Kingdom}
}

\renewcommand{\shortauthors}{Lee, Padh, Watson, Kilbertus, and Singh}

\begin{abstract}
  Fair machine learning (ML) methods support the identification and mitigation of the risk that algorithms will encode and\slash or automate social injustices. While algorithmic approaches alone cannot resolve structural inequalities, these techniques can support socio-technical decision systems by surfacing unintended discriminatory biases, clarifying trade-offs, and enabling governance.
  Although fairness has been well studied in supervised learning, many real-life ML applications are online and sequential, with feedback from previous decisions informing future decisions. Each decision in such a setting is taken under uncertainty due to unobserved counterfactual outcomes and finite samples, with especially dire consequences for under-represented groups, who are often systematically under-observed due to historical exclusion and selective feedback. For example, a bank cannot know whether a denied loan would have been repaid, and it may have less data about previously marginalized and financially excluded populations. 
  
  Towards this, this paper introduces a taxonomy of uncertainty in sequential decision-making---including \textit{model uncertainty}, \textit{feedback uncertainty}, and \textit{prediction uncertainty}---to provide a shared vocabulary for assessing and governing sequential decision systems in which uncertainty is unevenly distributed across groups. We formalize the model uncertainty and feedback uncertainty using counterfactual logic and reinforcement learning techniques. We illustrate the potential harms for both the decision maker (unrealized gains and losses) and the decision subject (compounding exclusion and reduced access) of na\"ive policies that ignore the unobserved space. We illustrate our framework using simple algorithmic examples that demonstrate the possibility of simultaneously reducing the variance in outcomes for historically disadvantaged groups while preserving institutional objectives (e.g. expected utility) of the decision maker. Our experiments on data, simulated to include varying degrees of bias, illustrate how unequal uncertainty and selective feedback can produce disparities in sequential decision systems, and how uncertainty-aware exploration can alter observed fairness metrics. By providing a structured lens for identifying where and how uncertainty arises, this framework equips researchers and practitioners to better diagnose, audit, and govern fairness risks in sequential decision systems.
  In sequential and online systems in which uncertainty is a core driver of unfair outcomes, not merely incidental noise, our results highlight the importance of explicitly accounting for uncertainty in fair and effective decision-making.
\end{abstract}

\maketitle

\section{Introduction}\label{sec:intro}
Machine learning (ML) algorithms can aid decision-makers in many high-stakes settings, such as healthcare, finance, and criminal justice. While resulting models are often faster and more accurate than human experts, they also pose an especially pernicious risk when historical inequities are reflected---not only in training data---but also from measurement, labelling, system design choices, and feedback from prior decisions \cite{barocas-hardt-narayanan,lee2021risk,corbett2018measure}.
For instance, if racial discrimination has artificially reduced the number of loans given to applicants from minority groups, then ML models may perpetuate this injustice at scale, automating harmful policies in a feedback loop that makes it difficult for qualified applicants from underrepresented minorities to access credit. To date, 
 this problem has been studied primarily in supervised learning. Academics have proposed a variety of (often mutually incompatible \citep{kleinberg2017}) fairness metrics for classification or regression models \citep{barocas-hardt-narayanan} and methods for imposing fairness constraints on the learning process itself \citep{zafar2019}. These fairness measurement and mitigation methods are best understood as tools to formalize particular values and trade-offs, rather than complete solutions, with their appropriateness depending on the socio-technical context. 
Existing fairness metrics designed for supervised models 
are ill-suited to many real-world ML applications if they do not capture how inequities compound over time under selective feedback and uncertainty -- our focus in this work. High-stakes settings often require dynamic decisions, rather than static predictions. In domains such as insurance pricing, fraud detection, hiring, and lending, predictions are not evaluated in a one-off mass dataset; rather, a decision is made for each individual or a batch of individuals, and the outcome of that decision informs future policies. In addition, each decision in such a setting is made under uncertainty, a subtlety that is generally overlooked in supervised ML.
For example, a bank cannot know whether a denied loan would have been repaid, and it may have less data about previously marginalized and financially excluded populations.
Such sequential decision-making procedures are more naturally modeled by reinforcement learning (RL) algorithms, which seek to maximize expected rewards over trials by exploring the unobserved space.\footnote{Though some prefer to reserve the term ``RL'' for policy optimization in complex dynamic systems with long-term horizons, we adopt a more inclusive usage that extends to simpler online learning settings such as bandits. This is common in standard textbooks, e.g. \citep{Sutton1998}.}

There has been a recent burst of interest in fair RL. (For survey introductions, see \citep{Zhang2021, gohar2025longterm, caton2024fairness})
While fairness and uncertainty considerations are well-studied in supervised learning (especially binary classification), comparable work in online settings is relatively limited. These gaps are discussed in \S\ref{sec:lit_rev} with recommendations for future research. Much of existing literature on RL has introduced several valuable technical methodologies, but their engagement with the nature or consequences of \textit{uncertainty} in fair decision-making in socio-technical systems, which is compounded by numerous observed and unobserved processes and competing incentives, has been limited. Unlike many of the technical papers in this area, %
here we do not introduce a novel fairness algorithm. Instead, we provide a taxonomy and diagnostic framework that gives practitioners and fair-ML researchers a shared vocabulary for identifying where uncertainty enters sequential decision pipelines, paving the way for more principled auditing, governance, and algorithm design in fair sequential decision-making. 
In our next section, we situate our work in the context of the debate around the legal and practical considerations of using RL in real-world settings.

\subsection{Legal, practical, and societal risks of RL} \label{sec:legal}

The use of RL in sequential decision-making can be challenging to justify to key stakeholders due to its potential risks. In this section, we briefly discuss the implications of RL beyond model performance metrics. RL carries legal risks if it is in violation of non-discrimination laws and regulations, reputational risks if misaligned to customer expectations, and practical risks if misaligned to business strategy. Recent work on the societal implications of RL deployment takes an important step in addressing this gap in literature \citep{khanna2025unveiling, whittlestone2021societal}, touching on a variety of topics (oversight, safety, reliability, and privacy) but with limited engagement on the alignment of algorithmic design with legal and regulatory constraints.

The RL algorithms and subsequent decisions can pose challenges with regards to non-discrimination laws. The international human rights legal framework, codified in the Universal Declaration of Human Rights and supported by other treaties and documents, establishes the principles of non-discrimination on the basis of certain features such as sex, race, language, or religion \citep{assembly1948universal}. These principles are reflected in different jurisdictions in local laws and regulations, such as the Equality Act in the U.K. \cite{equality-act-2010}, and in those targeting certain domains, such as fair lending laws in the U.S. \cite{fair-housing-act, ecoa-us}. 
While supervised ML metrics and mitigation techniques have been analyzed for their incompatibility with European non-discrimination law in particular \citep{wachter2021fairness}, limited attention has been given to the legal implications of RL. 

RL, however, has the potential, in many ways, to be \textit{more problematic regarding issues of discrimination than supervised ML from a legal standpoint}.
There are three contexts in particular in which RL can raise concerns with regard to certain non-discrimination laws: 1) in high-impact decisions, 2) when individual fairness is crucial, and 3) when there are few sequential decisions made in a limited time frame. RL works through exploration in a stochastic environment, necessarily introducing randomness into decisions. This can involve a nonzero probability of overstepping a legal, ethical or moral boundary,
e.g. denying a loan to someone who is expected to repay under the model. While this risk can be mitigated through both technical and process-driven (e.g. human-in-the-loop) guardrails, the residual risk of randomness may still persist in the system. %
As \citet{kilbertus2020} note in their article on fair RL, ``not all exploring policies may be (equally) acceptable to society''. The cost of exploration may be too high, unjustified, unjustifiable, or indeed, illegal in certain domain contexts, such as in criminal justice, employment, and essential financial services. 

Moreover, RL may not achieve individual-level fairness, which mandates that ``similar'' individuals are treated ``similarly'' \citep{dwork2012}. Traditionally, US non-discrimination laws and fair lending laws have scrutinized the \textit{inputs} of models, which may be effective for rules-based systems but less relevant for %
ML models that rely on complex correlations \citep{gillis2022input}. While past work on this topic focused on supervised ML, model inputs are even less informative for RL, in which stochastic learning may render individual decisions inconsistent, even with the exact same inputs. Indeed, even if RL would lead to a preferable policy at the group level, \textit{there may be suboptimal and potentially discriminatory decisions made in the exploration phase at the individual level, which can have direct legal consequences and liability implications}. A discussion of the gap between European non-discrimination laws and ML fairness metrics focuses on the importance of \textit{context} in determining legal liability, as well as the determination of a ``legitimate comparator'' that was unjustifiably treated better \cite{wachter2021fairness}. While further legal analysis is required beyond the scope of this paper, the introduction of randomness that RL entails makes it possible that two individuals with similar features obtain different outcomes, which, based on certain criteria \cite{wachter2021fairness}, could be in violation of European non-discrimination law. Further, due to the time taken for exploration and exploitation in RL, the optimal policy maximizing a target performance metric may not even be achieved in time using RL. For example, in \citet{kilbertus2020}, it is not until time step $t=50$ that stochastic strategies dominate the deterministic policy. 

In addition to these legal concerns, there may be reputational risks and business risks due to any misalignment between customer expectations and strategy. Even if an RL is compliant from a legal standpoint, customers may not perceive the process as fair if they disagree with the role of randomness in decision-making. Previous studies have found that there is variability in how people judge the fairness of different random procedures \cite{eliaz2014fairness}. Concerningly for RL, the extent to which randomness is perceived to be fair depends partially on how ``conventional'' or ``familiar'' the means of randomness are to people \cite{eliaz2014fairness}. Further work is needed to understand the technical and contextual conditions under which people may believe RL to be fair, both in order to inform policymaking around the use of RL in various domain areas and to help with algorithm design and communication strategy for those wishing to experiment with RL in decision-making. On the business or policy side, in the private or public sector, if the introduction of randomness exposes the decision-maker to intolerable risk, this may raise additional strategic, operational, and reputational harm.
 
Given the potential implications for RL, it may be tempting to assume that RL is unsuitable for practical applications despite its closer resemblance to real-world settings. However, this is arguably short-sighted given the potential for RL to better handle the uncertainties present in real-life decisions. In particular, RL explicitly models the exploration–exploitation trade-off and can adapt policies over time based on observed feedback, making it well-suited to settings where outcomes are partially observed and data is generated endogenously through decisions \cite{Sutton1998,whittlestone2021societal}.
In fact, we use uncertainty-aware exploration as an illustrative mechanism for examining how unequal uncertainty may influence fairness outcomes, in a way that is relevant to ongoing debates about the practical dimensions to fairness, including the relationship with non-discrimination law and regulatory oversight of automated decision systems. Our work was motivated by the need for approaches that %
both account for the unobserved decision space and %
attempts to accord with existing legal frameworks and customer expectations. However, further work is needed to better understand compliance requirements in RL algorithms for different jurisdictions and domain areas, which we flag as an explicit area for future work in \S\ref{sec:lit_rev}.
The above considerations highlight the governance challenges associated with deploying sequential learning systems in high-stakes domains. Our taxonomy in \S\ref{sec:taxonomy} is intended to help practitioners identify where uncertainty arises in such systems and to inform auditing and oversight processes.

\subsection{Towards fairness under uncertainty for RL}
Uncertainty is more harmful to some groups than to others. In some settings, this may arise through selective surveillance and institutional scrutiny. In other settings, marginalized populations will often have less data (e.g., due to lack of financial inclusion in lending), and therefore predictions about them will be made with less confidence. Our focus is on the latter. 
This is a known bias, often termed ``representation bias'' \citep{lee2021risk}, which can inflate the risk profile of marginalized groups, negatively affecting both profitability and equitable distribution of financial opportunities. 

Within our taxonomy, this dynamic arises from the interaction between prediction uncertainty and feedback uncertainty across groups. To illustrate this mechanism, we consider a simple form of targeted exploration that accounts for differences in group-level uncertainty. Intuitively, someone in a minority group should not be penalized for the model's uncertainty, especially if the uncertainty is due to a history of exclusion, discrimination, or marginalization. We therefore consider decision policies that allocate exploration in proportion to uncertainty, for example by increasing the likelihood of granting a loan when prediction uncertainty is high. This is meaningfully different from affirmative action, which (controversially) favors minority groups \citep{garrison2004affirmative}. Rather than introducing group-based preferences, the approach highlights how uncertainty itself can act as a structural source of disparity, and how accounting for it can alter decision outcomes. IWe describe this illustrative mechanism in more detail in \S\ref{sec:theory}. The \textit{exploration–exploitation trade-off}, a fundamental concept in RL, provides a useful lens for understanding how such policies balance information gathering with performance optimization.

Our illustrative exploration strategy serves as a concrete instantiation of the framework, demonstrating how the uncertainty categories identified in our taxonomy, particularly prediction and feedback uncertainty, interact in sequential decision settings, and how resulting differences in group-level uncertainty can shape exploration–exploitation trade-offs with implications for legal and governance considerations. 
This is an \textit{active acknowledgment of uncertainty in the ML system}, challenging the decision-maker to consider the known unknowns. The amount of exploration should be proportional to the amount of uncertainty and the decision-maker's risk appetite. It should not be na\"ively assumed that all denied loans would have defaulted, as this assumption would especially harm marginalized groups. Instead, the assumptions around the current decision boundary should be constantly challenged, taking seriously the differences in uncertainty the decision-maker has about each individual and each sub-group.

\looseness=-1
Our main contributions are: 
\begin{enumerate}
\item We introduce a taxonomy of uncertainty in sequential decision systems, bringing together different literature to provide a holistic and structured vocabulary for identifying and reasoning about how different forms of uncertainty arise and interact across the ML lifecycle;
\item %
Using this taxonomy, we formalize the problem of ``fairness under uncertainty'' for RL in a socio-technical system, conceptualizing how unequal epistemic and feedback uncertainty can compound disparities under selective labels, and 
\item We implement simulations that demonstrate the consequences of the mechanisms identified in our taxonomy, showing that even simple uncertainty-aware exploration can improve fairness metrics without explicit fairness constraints and without sacrificing utility, providing an illustration of the framework's diagnostic value. 
\end{enumerate}

The remainder of this paper is structured as follows. In \S\ref{sec:taxonomy}, we introduce our taxonomy of uncertainty in sequential decision-making. We formalize our problem setup in \S\ref{sec:theory}, where we use counterfactual logic to demonstrate the biases that such uncertainty can engender.
We describe our data in \S\ref{sec:method} and our proposed method in \S\ref{sec:method}. Experimental results follow in \S\ref{sec:exp}.
We review the literature on fair RL in \S\ref{sec:lit_rev}, highlighting the key gaps we address in this work. We propose a practical solution illustrating its performance on a range of benchmarks. Following a brief discussion on limitations and future work, \S\ref{sec:concl} concludes.

\section{Taxonomy of uncertainty}\label{sec:taxonomy}
Scholars have studied subsets of various uncertainties in the ML lifecycle, e.g. epistemic and aleatoric uncertainty \cite{hullermeier2021aleatoric}, predictive uncertainty in fairness evaluation \cite{romano2020malice}, and robustness to distributional shifts \cite{bhatt2021uncertainty}. However, these works typically analyze uncertainty at specific stages of the ML pipeline.
Our taxonomy organizes these sources of uncertainty that have been previously studied across the ML lifecycle and highlights their interactions in sequential decision systems. From a governance perspective, the taxonomy provides a structured way to audit where uncertainty enters a decision pipeline, from data collection and feature construction to prediction and feedback. Identifying these sources of uncertainty can support risk assessment, model evaluation, and documentation practices.

\begin{figure}[h]
  \begin{center}
    \includegraphics[width=0.6\columnwidth]{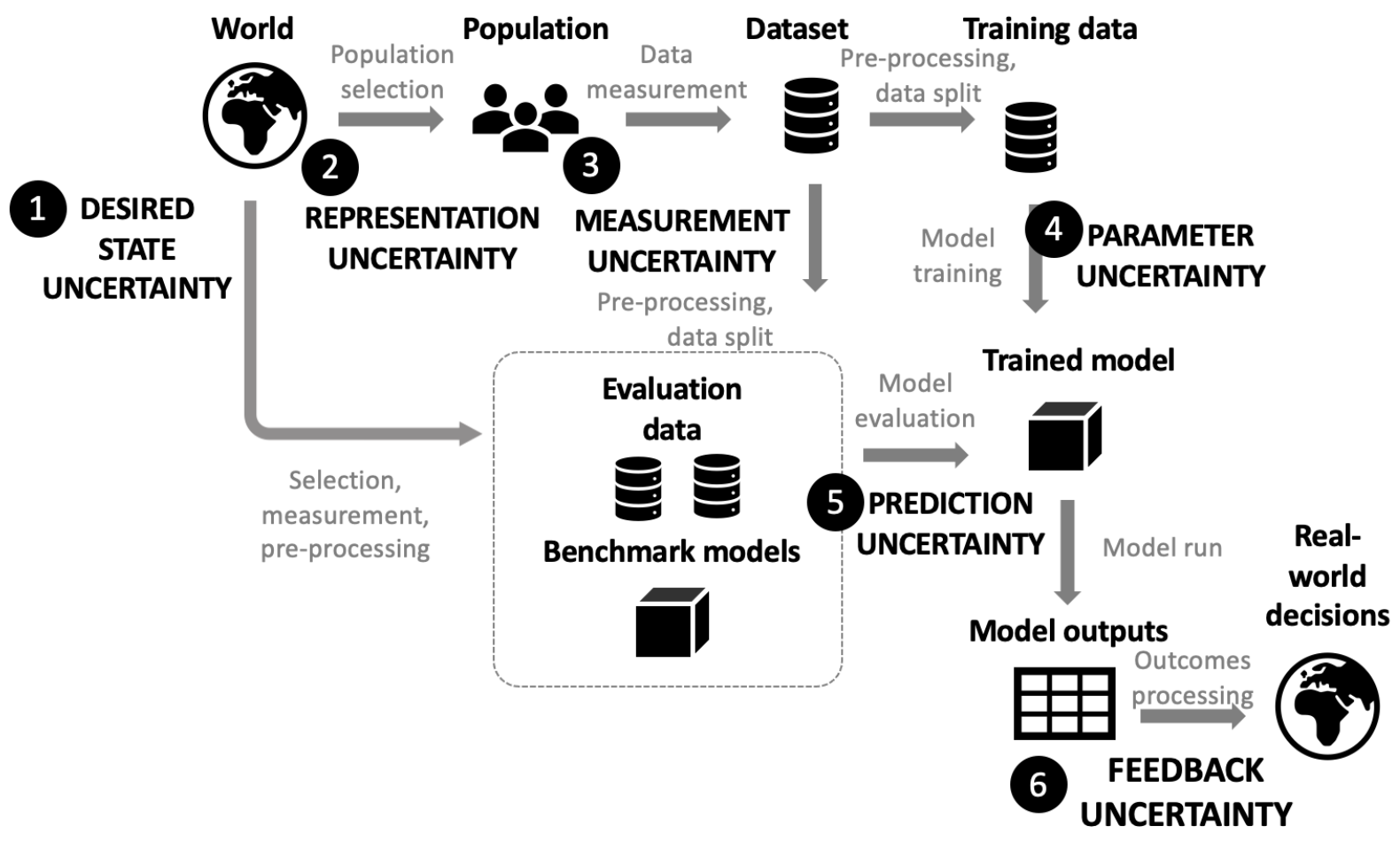}
  \end{center}
  \caption{Uncertainty taxonomy throughout ML lifecycle}
  \label{fig:taxonomy}
\end{figure}

Figure ~\ref{fig:taxonomy} shows the six types of uncertainty we consider. This is derived from a review of ML literature discussing sources of uncertainty, discussed below, which we mapped to the different phases of a typical ML lifecycle. While this forms only a part of the contribution of this paper, it provides structured vocabulary for communicating past related work and for analyzing uncertainty-driven harms in sequential decision systems. The taxonomy provides a conceptual lens for understanding how uncertainty arises at different stages of the ML lifecycle and how these uncertainties interact in sequential decision systems. By organizing these mechanisms into a unified framework, we aim to support more systematic analysis of uncertainty-driven harms. The taxonomy can help different stakeholders identify where disparities arise, e.g., enabling practitioners to diagnose issues in data or model design, researchers to study the interaction of prediction and feedback uncertainty, and auditors or regulators to assess the role of selective feedback and uneven uncertainty across groups.
Similar visualizations of the ML lifecycle have been used in previous work \cite{lee2021landscape}. Uncertainties 1-4 are ``global'' uncertainties that affect the model on a systemic level. These should be considered to inform design choices throughout the ML lifecycle. 5 and 6 are ``local'' uncertainties on an individual or sub-group level, which should be implemented to actively take into account in the ML algorithm. In practice, these distinctions can guide auditing processes, e.g., by distinguishing uncertainties that should be addressed through data collection and institutional design (global uncertainties) from those that arise during model deployment and require monitoring or exploration policies (local uncertainties).

The probabilistic ML literature has distinguished between \textit{aleatoric} and \textit{epistemic} uncertainties \cite{hullermeier2021aleatoric, bhatt2021uncertainty, romano2020malice}. In the context of ML, aleatoric uncertainty refers to inherent stochastic variability, such as that of a (fair) coin toss. Epistemic uncertainty, by contrast, is due to a lack of data and tends toward zero as sample size increases. The former gives information on noise or class overlap of the data, while the latter shows where performance can be improved by collecting more data in input regions where the training dataset was sparse \cite{bhatt2021uncertainty}. Recent work has also explored the relationship between uncertainty and fairness. For example, Romano et al. (2020) \cite{romano2020malice} propose uncertainty-aware coverage guarantees across groups, while Kuzucu et al. (2024) \cite{kuzucu2024uncertainty} frame fairness in terms of differential susceptibility to noise or missing data. These approaches highlight how unequal uncertainty can affect predictive performance across groups but largely focus on static supervised learning settings.

While this framing of reducible (epistemic) and irreducible (aleatoric) uncertainties is useful, there are various sources of both types throughout the ML development process. Our taxonomy builds on this literature by situating different uncertainty types across the ML lifecycle and highlighting how they interact in sequential decision settings. This section aims to break down each uncertainty into its component sources. 

\textbf{1. World: Desired state uncertainty~} The fundamental uncertainty on the level of worldview is: how much of the existing inequality is undesirable and should be actively corrected? For example, in lending, discrimination in the job market may increase the actual credit risk of women compared to men. \citet{lee2021formalising} argue that it is up to the decision-maker to define which types of inequalities are acceptable or unacceptable in any use case, including inequalities in genetics, talent, and socioeconomic ability. To an extent, this is a subjective judgement without a single answer. However, the uncertainty can be reduced through a better understanding of types of inequalities and their impact on the decision-maker's key objectives. Therefore, it can be classified as a type of epistemic uncertainty.

\textbf{2. Data collection: representation uncertainty~} To what extent is the dataset skewed compared to the target population? ``Representation bias'' is a known issue among industry practitioners, which is the skewed sampling of the training data \cite{lee2021landscape, holstein2019improving}. Representation \textit{uncertainty} refers to the unknown variability of the size and direction(s) of this bias and is a type of reducible (epistemic) uncertainty. The known mitigation technique is to test for representation (e.g. proportion of women in dataset) against a known population (e.g. proportion of women in the country) and consider additional data collection or reweighting \cite{lee2021risk}. This is closely related to literature on selective labels (e.g. \cite{lakkaraju2017}), which examines how decision-dependent data collection can bias training data. These works focus primarily on estimation and evaluation challenges under selective observation. Our taxonomy complements this literature by highlighting how uncertainty generated by selective feedback interacts with other forms of uncertainty across the ML lifecycle.

\textbf{3. Feature engineering and selection: measurement uncertainty~}
How well do our features measure what the decision-maker would like to measure? It is well understood that datasets may contain undesirable proxies of demographic variables \cite{corbett2018measure}. This type of bias (measurement bias) embedded in features may arise due to issues of data quality, for which there have been several techniques proposed for practical assessment and mitigation \cite{loshin2010practitioner}. This is a type of reducible (epistemic) uncertainty, as it represents the gap between what is known and what is true.

\textbf{4. Model training and build – model uncertainty~}
How close are the model parameters to a ``true'' model? The model built may not be the optimal policy, and a model may exist that better represents the feature relationships. This is a type of reducible (epistemic) uncertainty but differs from measurement uncertainty in its focus on the model type and parameters rather than its training features. \citet{bhatt2021uncertainty} breaks this down further into \textit{model uncertainty} in reference to model parameters and \textit{model specification uncertainty} in reference to the model type. We group them together here because they are a part of the same phase of the ML lifecycle in training and building a model. \citet{Dimitrakakis2019} addresses this uncertainty through RL. However, their formalization does not address the remaining uncertainty, and it is unclear how they are compatible with non-discrimination laws, as discussed in \S\ref{sec:intro}. Our proposed approach addresses the model uncertainty through RL that explores the unobserved space in a targeted way that takes into account the decision-maker's risk appetite.

\textbf{5. Predictions and test data: prediction uncertainty~}
The uncertainty around each prediction (e.g. of probability of repayment) may vary by applicant. There has been work around confidence intervals for each prediction based on the training data, but rarely in the context of fairness \citep{romano2020malice}. In particular, uncertainty for previously marginalized or excluded groups may be comparatively high, due to representation bias \citep{lee2021risk}. This is a key uncertainty we aim to address, by comparing decision boundaries between individuals, to be discussed in \S\ref{sec:theory}. At a point in time, this is a type of aleatoric (irreducible) uncertainty because for each applicant there is a distribution of plausible outcomes. However, on a systemic level in an online learning setting, it can also be considered epistemic (reducible) because as the model gains more data about similar applicants, its uncertainty around any particular applicant should decrease over time.

\textbf{6. Deployment and retraining – feedback uncertainty~} 
Often, the decision-maker’s actions determine what data are collected. For example, a rejected job applicant’s performance is not measured, and whether a denied loan would have been repaid is unknown. A targeted RL can explore whether the decisions were valid through an analysis of the counterfactual: the expected utility had the decision been different. This uncertainty only exists in online learning settings due to the ``known unknowns'' in the system. This is slightly more difficult to classify, as it is an epistemic uncertainty in a sense that it is due to the modeller's limited knowledge about the state (or in this case, the counterfactual state) of the world. However, while epistemic uncertainty can always be reduced by collecting more data, in this case, the counterfactual state is unobservable in principle. We formalize this point in \S\ref{sec:theory}. 

While this paper will directly address uncertainties 4-6 (model uncertainty, prediction uncertainty, and feedback uncertainty), it is important to contextualize the algorithm in the global uncertainties that exist at a systemic level. For example, if there are doubts associated with the training data quality 
and representativeness, the reliability of the model prediction should also be in question. Each of the global uncertainties should be mitigated at their source: desired state uncertainty by discussion with key stakeholders, representation uncertainty by collecting more data, and measurement uncertainty by fixing data quality issues. These mitigation strategies have been well understood and studied in their related bodies of literature. In the remainder of the paper, we illustrate how three types of uncertainty from the taxonomy (model uncertainty, prediction uncertainty, and feedback uncertainty) can interact under selective labels to produce disparities in sequential decision settings. We focus on these three because they operate directly at the decision stage and are most immediately relevant for sequential learning dynamics. While the taxonomy includes additional sources of uncertainty across the ML lifecycle, this subset captures the core mechanisms through which unequal uncertainty can propagate under selective feedback. The analysis is therefore illustrative rather than exhaustive, demonstrating how the framework can be applied in practice.
In the next section, we introduce the theory behind our approach in mitigating these uncertainties. 

\section{Uncertainty mitigation}
\label{sec:theory}

The three key uncertainties in the scope of our paper are: model uncertainty, prediction uncertainty, and feedback uncertainty. 
Model uncertainty is global: on a system-level, there is uncertainty about whether the current model is optimal. It may result in better outcomes by shifting the existing decision boundary for those who receive a loan. This is compounded by a related local uncertainty (feedback uncertainty): the decision-maker can only observe the true outcomes of loans that have been approved. These two uncertainties are usually tackled through targeted reinforcement learning. Prior work by Kilbertus et al. (2020) defined ``semi-logistic policies,'' which combines deterministic and stochastic approaches: his policies deterministically approve examples believed to contribute positively to the utility by the current model and only explore stochastically on the remaining ones, which was designed to address representation uncertainty in particular. Our work builds on this approach to diagnose how unequal uncertainty can structurally generate disparities in such sequential decision systems, situating these mechanisms within a broader taxonomy of uncertainty in sequential decision systems.

The intuition behind this semi-logistic decision-making process is to use supervised learning to make decisions for the ``clear-cut'' cases and to take calculated risks on the uncertain cases. Those predicted to repay with high probability would receive a loan, and those with low probability of repayment would not receive a loan. The size of the ``uncertainty boundary'' may depend on the decision-maker's risk appetite, the amount of global uncertainty as discussed in the taxonomy (\S\ref{sec:taxonomy}) and the amount of historical data available. This allows the decision-maker to explore whether the current model is optimal, taking into account the unobserved space.

Prediction uncertainty is another local uncertainty in scope. %
Even though two individuals may have similar predicted probability of repayment, they may have different uncertainties around the predictions. In particular, those from previously excluded and marginalized groups may have wider probability distributions due to limited representation of their success in the datasets. For example, if Applicant A, who is from a privileged group, has a probability distribution that is slightly higher but overlaps with Applicant B, who is from an under-privileged group, then there is a probability range in which Applicant B outperforms Applicant A. As an illustrative example, we consider a simple adjustment in which the model probabilistically explores decisions when prediction uncertainty overlaps across groups, so that the model takes a chance on Applicant B in proportion to the probability that he or she would outperform Applicant A. 

In this section, we first formalize the utility calculation that considers both observed outcomes and unobserved outcomes, tackling the feedback uncertainty. We propose a contextual bandit algorithm 
designed to challenge existing decision boundaries, in order to illustrate how model uncertainty manifests at the decision boundary and how targeted exploration can reveal and reduce this uncertainty over time. Finally, we propose a policy that does not penalize under-privileged applicants for a model's uncertainty about their predictions. A key point here is that we do not explicitly model fairness constraints in our implementation. Rather, we only account for the uncertainties that are practically actionable to mitigate at the model level, which we illustrate can improve fairness outcomes.

\textbf{Setup. } At each time point $t \in \mathbb{N}$, we see a new applicant with observable data $\bm{x}_t$, including sensitive attribute $A \in \{0, 1\}$, where $A=1$ denotes membership in the advantaged majority and $A=0$ membership in the disadvantaged minority. Assume a state space $\mathcal{S} = \mathcal{X} \times \mathcal G \times \mathcal L \times \mathcal{C}$, comprising applicant profiles $\mathcal X$; gains $\mathcal G$ and losses $\mathcal L$ associated with a given loan; as well as the bank's constraints $\mathcal C$ (e.g., regulatory influences and budget).
We do not observe the potential outcome variable $Y_1 \in \{0, 1\}$, which indicates whether the customer would repay the loan if given the chance---unless, of course, the bank decides to make the loan, in which case we do observe $Y_1$. We assume that this depends on the applicant profile $\bm X$, as well as the size of the loan, parameterized by $G,L$. Let $D \in \{0, 1\}$ denote the decision variable, governed by a policy $\pi: \mathcal{S} \mapsto [0, 1]$, which evolves over time. 

Conventional methods focus exclusively on \textit{observed} profits; however, this ignores \textit{unrealized} gains and losses, which may be substantial.  Failure to grant loans to applicants who would repay incurs an opportunity cost, while failure to grant loans to applicants who would default represents sound judgment. Ignoring these outcomes---or worse, assuming that all denied loans would have resulted in default---creates pernicious feedback effects that are detrimental to both customers and lenders. 
The true utility of a given policy must be calculated over factual and counterfactual instances:
\begin{align*}
    u &:= u(d, y_1, g, l)\\
    &= \underbrace{d(y_{1} g - (1 - y_{1}) l)}_\text{observed} +
    \underbrace{(1 - d) ((1 - y_{1}) g - y_{1} l).}_{\text{unobserved}}
\end{align*}
The reward function for policy $\pi(\bm s) := \argmax_d ~p(d \mid \bm s)$ is the expected utility as states and potential outcomes vary:
\begin{align*}
    r(\pi)
    &= \sum_{\bm s, y_1} ~p(\bm s) ~p(y_1 \mid \bm s) ~u\big(\pi(\bm s), y_1, g, l\big),
\end{align*}
with the understanding that $g, l$ are determined by $\bm s$.
The optimal model $\pi^*$ from a set of candidates $\Pi$ is whichever maximizes this quantity:
\begin{align*}
    \pi^* := \argmax_{\pi \in \Pi} ~r(\pi).
\end{align*}
In the following subsections, we discuss how we tackle each of feedback uncertainty, prediction uncertainty and model uncertainty so as to maximize expected utility. It is important to note, however, that these uncertainties are not independent but can be related to each other, and this will also be reflected in our suggestions for mitigation.

\textbf{Feedback uncertainty.}
\label{sec:feedback}
The myopic focus on the first row of Table \ref{tab:conf_mat} ignores the opportunity cost of missing out on customers who would, in fact, pay back their loans. Assuming that all $D = 0$ cases are dodged bullets introduces bias. Note that, with access to unobservables, this policy is trivially identified as $D = Y_1$. In practice, we must employ a model to estimate the potential outcome, $p(Y_1 \mid \bm{x})$.\footnote{Readers more familiar with Pearl's do-calculus \citep{pearl2009causality} may prefer the notation $p\big(Y \mid \bm x, do(D=1)\big)$, which emphasizes that our target is a conditional causal effect.} Assuming that $D$ depends on $\bm{X}$, this function can only be learned from experimental data. 
One approach is to divide all data into two regimes: observational $(F_0)$ and interventional $(F_1)$. With probability $\delta>0$, subjects are recruited for an experiment (assignment mechanisms detailed below). If recruited, loans are provided independent of applicant profiles $\bm{X}$. Data from these trials is used to estimate $p(Y_1 \mid \bm{x})$. 

\begin{table}
\caption{Confusion matrix of decisions and potential outcomes. Only the top row is observed by the agent.}
\label{tab:conf_mat}
\small
\setlength{\tabcolsep}{4pt}
\begin{tabular}{ccc}
& \multicolumn{2}{c}{Potential Outcome, $Y_1$} \\
\cmidrule{2-3}
Decision, $D$ & 1 & 0 \\ \midrule
\multicolumn{1}{c|}{1} & realized gain & realized loss \\
\multicolumn{1}{c|}{0} & unrealized gain & unrealized loss \\
\hline
\end{tabular}
\vspace{-3mm}
\end{table}

\textbf{Prediction uncertainty.}
\label{sec:prediction}
In a loan setting, groups of individuals who are historically disadvantaged would have a lower representation in the data and, therefore, a higher degree of uncertainty in their predictions. There are various ways that prediction uncertainty could be mitigated. A direct way is to get confidence intervals around each prediction and perform an uncertainty-aware exploration based on the intervals, as in the well known UCB algorithm. This could be achieved either by randomly exploring among the applicants whose confidence intervals overlap with the decision threshold, or a targeted exploration of individuals for whom the interval width is the widest.
If \(\hat{p}_\text{repay} \pm \Delta\) represents the predicted probability of repayment and its associated uncertainty, the decision boundary can be adjusted as follows for instance:
\[
D_\text{adjusted} = 
\begin{cases} 
1 & \text{if } \hat{p}_\text{repay} + \Delta > \text{threshold}, \\
0 & \text{otherwise}.
\end{cases}
\]

\textbf{Model uncertainty.}
\label{sec:model}
Model uncertainty can manifest globally, as uncertainty about the optimality of the model's decision boundary, or locally, as uncertainty in underrepresented regions of the feature space. We define model uncertainty as the gap between the expected utility of the current model and the utility of the optimal model. Mathematically, this is expressed as: $\mathcal{R}(\pi) = r(\pi^*) - r(\pi).$
This formalization illustrates how model uncertainty can be conceptualized in sequential decision settings, in which the goal is to minimize this value with high probability.

There are several practical ways to mitigate model uncertainty. One approach is to compare observed gains (\(\mathbb{E}[u_t^{\text{obs}}]\)) with unrealized gains (\(\mathbb{E}[u_t^{\text{unobs}}]\)) to assess the opportunity cost of unobserved decisions. Additionally, the variance in utility (\(\text{Var}(U|a)\)) can be used to highlight regions of high variability in observed profits, indicating potential uncertainty. Finally, uncertainty near the decision boundary—where predictions lack confidence—serves as a key area for exploration to refine the model and improve decision-making. Bayesian optimization and bandit algorithms are other direct alternatives, as they can explicitly model the uncertainty of the under-observed space \citep{kandasamy2015high, tuo2022uncertainty}.

\section{Methodology}\label{sec:method}

\paragraph{Data description} We use the domain area of loan applications for our illustrative example. We use a pipeline proposed by Baumann et al. \cite{baumann2023bias} to simulate different types and magnitudes of bias in binary decision problems. Specifically, their framework accommodates historical bias, measurement bias, representation bias, and omitted variable bias through a set of structural equations that users can tune in a simple, unified interface. Further details are provided in \cref{app:data}. We used synthetic datasets in order to evaluate counterfactuals, which are otherwise unobservable by definition, and to
retain control over different types and magnitudes of bias to understand the potential impact of different methods on bias and performance. Existing datasets commonly used in the Fair ML literature, such as Adult and German Credit, would not have allowed for this. In addition, these datasets do not have the temporal element necessary for RL settings.

\textbf{Debiasing approaches} Having described the general ways in which each of the uncertainties can be mitigated, we will describe our approaches. The bias mitigation strategies we use are deliberately simple, as their purpose is to make the mechanisms identified in our framework transparent. By isolating how different forms of uncertainty interact in sequential settings, these simulations illustrate how even minimal uncertainty-aware adjustments can shift fairness outcomes without sacrificing utility. They are not intended as fairness guarantees or production-ready solutions, but rather as a foundation for developing more robust approaches.

All methods use logistic regression as the base model, which, while simple and interpretable, successfully illustrates the possibility of enhancing outcomes by accounting for uncertainty. 
This has the added bonus of providing parametric confidence intervals for all predictions. We compute confidence intervals via the Delta Method applied to the predicted probabilities: $\Delta_i = z_{1-\alpha/2} \cdot \sqrt{\nabla p_i^\top (X^\top W X)^{-1} \nabla p_i}$, where $W = \mathrm{diag}(p_i(1-p_i))$ is the weight matrix and $\nabla p_i = p_i(1-p_i)\bm{x}_i$ is the gradient of the sigmoid with respect to the linear predictor. We use 95\% confidence intervals throughout ($z_{0.975} = 1.96$), so $\Delta_i$ in \cref{sec:prediction} is this half-width.
In principle, the logistic regression model could be replaced with any function that provides probabilities for class labels, but may require nonparametric confidence intervals, which will be wider than the parametric alternative when model assumptions are (approximately) satisfied. Thus, these models illustrate that even taking simple uncertainty mitigation measures can help improve fairness without loss of utility.

A commonality between all except the counterfactual methods is that in each round of decision making, which in our case translates to each calendar quarter, we select a subset $\delta$ for exploration, as described in \cref{sec:feedback}. The differences are in how this subset $\delta$ is chosen. It is important to note here that we perform this exploration only on the subset of applicants which lie in the `uncertain region,' which is the region between a clear acceptance and rejection threshold set for the probability of repayment. This region---a global uncertainty---can be defined by a combination of domain expertise and the decision maker's risk appetite. 

Compared to the description in \cref{sec:feedback} we simplify the setting even further by not having a separate estimate of $p(Y_1 \mid \bm{x})$ from the exploration subset, but instead simply training on the payment feedback received from all applicants granted the loan. We use the first quarter (Q1) as a base quarter in which we have all the ground truth available, to be able to train a base model which is then updated sequentially each quarter (Q2-Q10).

\textbf{Na\"ive exploration.}
A random subset of applicants in the uncertain region is selected for exploration, such that any applicant could be chosen with probability $\delta$. Everyone in this exploration set gets a loan.
The remaining applicants are assessed according to a logistic model that has been trained up to the previous quarter. 
Accept/reject decisions are based on hard thresholds that can be tuned across runs.
Rather than building a new model each quarter, we update the same logistic model based on the feedback received from all approved loans---including those from the exploration set. This remains true for the subsequent methods as well.

\textbf{Weighted exploration.}
We take a stratified sample of applicants from the uncertain region weighed by their probability of repayment as calculated by the current logistic model up to that point. We continue sampling until the budget runs out, and this set is used for exploration.

\textbf{Uncertainty-aware exploration.}
We restrict our uncertainty region to applicants for whom confidence intervals overlap with the decision threshold. This method therefore explicitly takes prediction uncertainty. From this restricted uncertainty region, we sample randomly until the budget runs out. 

\textbf{Counterfactual utility maximization.}
We train a single logistic regression model end-to-end by maximizing a differentiable combined observed and counterfactual utility objective. Concretely, for each applicant $t$ with predicted repayment probability $p_t$, we define a soft decision $D_t = \sigma(\alpha(p_t - \tau_t))$, where $\sigma$ is the logistic function and $\alpha$ is a sharpness parameter controlling how closely the soft decision approximates a hard threshold. The training objective is: $\max_\theta \sum_t \Bigl[ D_t(Y_t g - (1-Y_t)l) + (1-D_t)(p_t g - (1-p_t)l) \Bigr]$, where the second term is the expected counterfactual utility for applicants not granted a loan. Using $\sigma(\cdot)$ for $D_t$ makes this objective differentiable, enabling gradient-based optimization. The threshold $\tau_t$ is set dynamically each quarter based on the available budget. This approach thus explicitly accounts for feedback uncertainty via the counterfactual term, without requiring a separate exploration model, while also accounting for a fixed budget.

The five methods form a structured ablation where each comparison isolates a distinct uncertainty mechanism. Moving from Naïve to Naïve Exploration isolates the effect of exploration alone on feedback uncertainty, while the shift to Uncertainty-Aware or Prob-Weighted variants isolates prediction uncertainty and probability-prioritized exploration respectively. Finally, the Counterfactual Utility method isolates direct feedback uncertainty mitigation through end-to-end utility optimization over both observed and counterfactual outcomes

\section{Experiments}
\label{sec:exp}

\looseness=-1
We use the synthetic data generation as described in \cref{sec:method}, simulating different magnitudes and types of bias which have been described, and see how out mitigation methods perform in these different settings. These simulations are designed to illustrate the mechanisms identified in the taxonomy rather than to benchmark deployable fairness algorithms.

\looseness=-1
$A$ is the binary sensitive attribute, with $A=0$ representing the minority group. Selection rate refers to the proportion of candidates from a specific group who are given a loan, with selection rate difference then being the difference in rate selection between the two sensitive groups. Other fairness metrics are more standard, and defined in \cref{app:evaluation}.

\begin{figure}[t]
    \centering
    \begin{subfigure}[b]{0.9\textwidth}
        \centering
        \includegraphics[width=\textwidth]{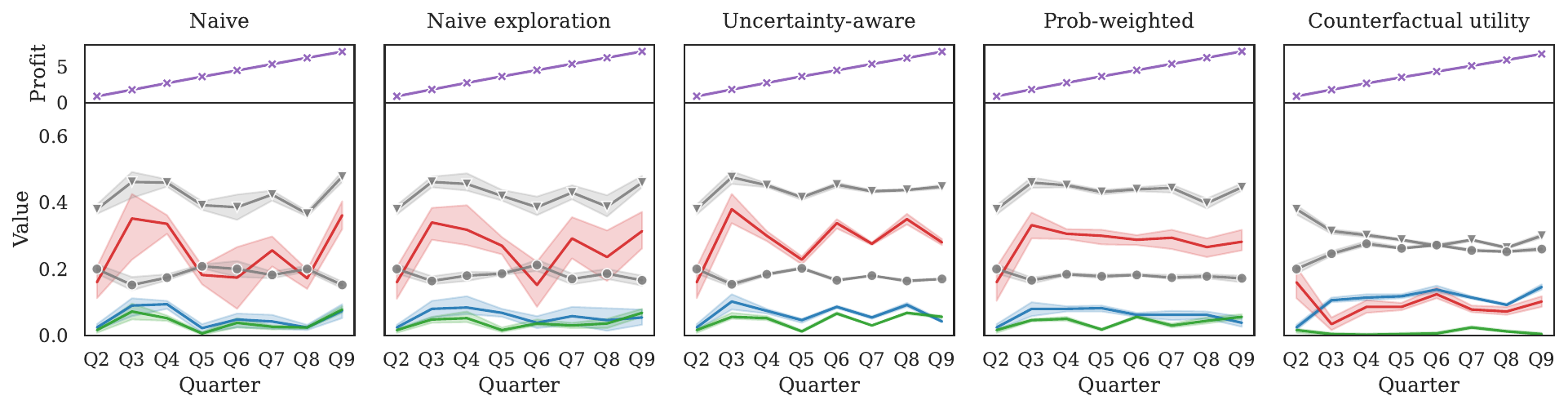}
        \label{fig:y-bias-met}
    \end{subfigure}
    
    \vspace{1em} %

    \begin{subfigure}[b]{0.9\textwidth}
        \centering
        \includegraphics[width=\textwidth]{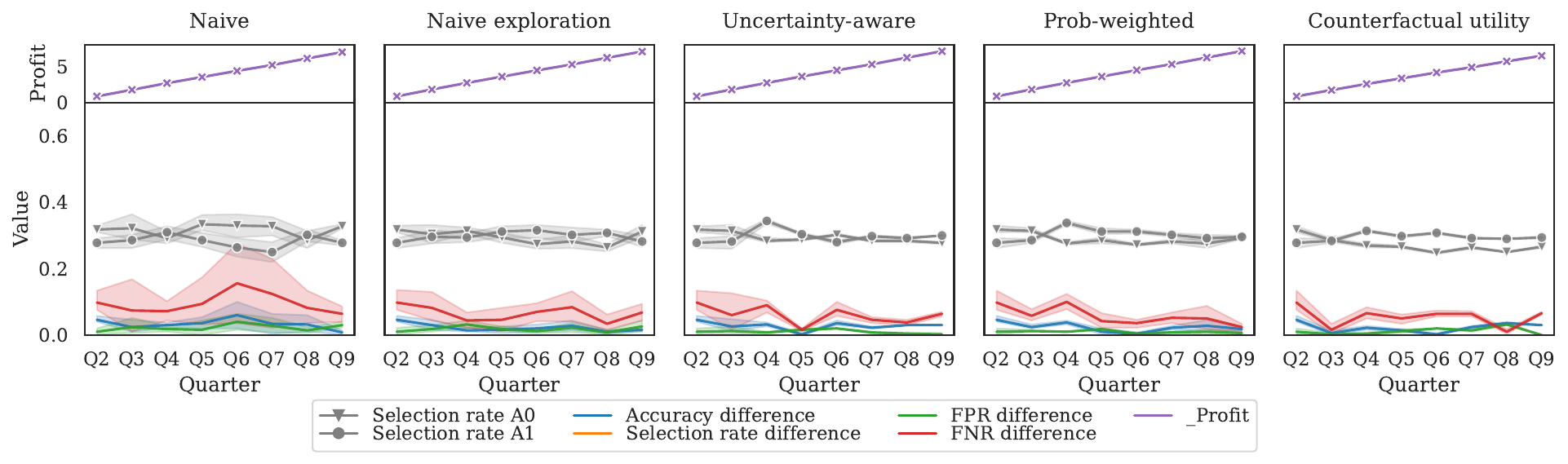}
        \label{fig:no-bias-met}
    \end{subfigure}

    \caption{Comparison of methods on performance and bias over time. The \textbf{top} figure is a setting with historical and measurement bias on the outcome Y. The \textbf{bottom} figure is a baseline setting with no bias. All methods perform similarly, confirming that mitigation strategies do not harm performance when bias is absent.}
    \label{fig:compare-methods}
\end{figure}

Fig. 2 compares the different methods and plots the evaluations of the performance and bias metrics across the different quarters in sequence. In our simulation, for moderate levels of historical and measurement bias, counterfactual utility maximization is much better at mitigating bias without a cost of utility (cumulative profit). While we cannot make generalized statements about its effectiveness across scenarios, this illustrates how accounting for unrealized outcomes can shift observed fairness metrics. Comparing the methods for different bias levels, we see that counterfactual utility is able to account for historical and measurement bias in $Y$ by explicitly looking at the unrealized and realized outcomes instead of a myopic focus on realized outcomes.

\begin{figure}[t]
    \centering
    \begin{subfigure}[b]{0.9\textwidth}
        \centering
        \includegraphics[width=\textwidth]{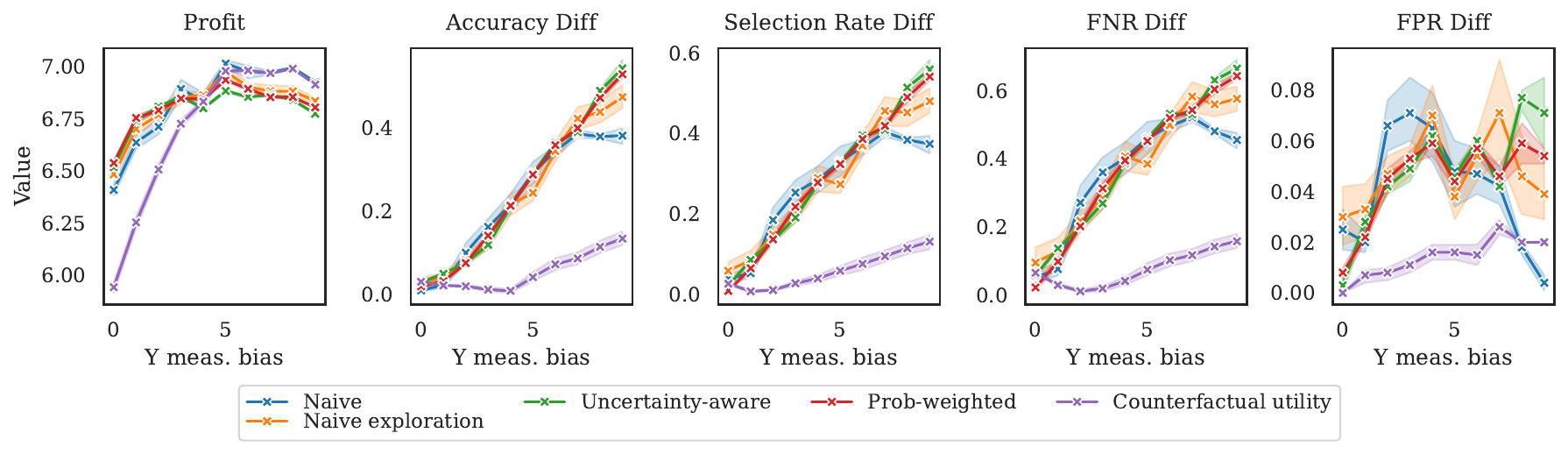}
        
        \label{fig:y-meas-bias}
    \end{subfigure}
    \caption{Sensitivity to $Y$ measurement bias at end of Q10, keeping the other biases at $0$.}
\end{figure}
Fig. 3 compares only the final values of the metrics at the end of our given time period (Q1-Q10). As we increase the magnitude of a specific bias metric, keeping the others fixed, we are able to obtain both better bias and better utility values in counterfactual utility maximization, without putting any explicit fairness constraints in the optimization but only accounting for uncertainty. With increasing levels of measurement bias on $Y$, the counterfactual utility maximization gives us a better utility as well as much better fairness values. The other methods are seen to be very similar. Note that the FPR parity looks noisy due to the small scale. While the initial values of the profit may seem significant, the scale only spans from 6 to 7. While it is expected that the profit is lower for utility maximization when there is no bias, it is noteworthy that this disparity disappears very quickly even for low levels of measurement bias on $Y$.

In summary, optimizing counterfactual utility promotes better outcomes under various fairness measures, at little to no cost in overall profits. Results vary depending on the source and magnitude of the underlying bias, but our experiments illustrate how ignoring the uncertainty arising from unrealized decisions can incur a penalty on both lenders and customers in certain settings when operating in environments with non-zero bias.

\section{Discussion and related work}\label{sec:lit_rev}
The literature on fairness in supervised learning is vast and heavily focused on defining metrics for model auditing (in the post hoc case) or proposing new constrained optimization techniques (in the ex ante case). Familiar examples of the former include demographic parity and equality of opportunity \citep{hardt2016, berk2001}, while the latter includes various methods for encoding such desiderata into the empirical risk minimization process itself \citep{donini_fair_erm, zafar2019}. A recent survey of ML fairness metrics and mitigation techniques \citep{caton2024fairness} provides an overview of different approaches, while acknowledging that much of the literature is focused on binary classification. 

Fairness measures typically impose some (conditional) independence relation that is evaluated with respect to its first moment. For example, equality of opportunity is satisfied for values $a, a'$ of some sensitive attribute $A$ when $\mathbb{E}[\hat{Y} \mid a, y] = \mathbb{E}[\hat{Y} \mid a', y]$, where $Y$ and $\hat{Y}$ denote true and predicted outcomes, respectively. Such measures ignore the differential impact of inequalities between higher moments of the conditional distribution. When the uncertainty of a predicted outcome is high, risk-averse agents (e.g.,  banks) may rationally prioritize decisions about which they feel more confident, even if the average payoff is equal. There has been acknowledgment of this issue in supervised learning settings. One study demonstrated that if the input is not represented in the training data set (``atypicality'') of a supervised binary classification model, this may result in poor accuracy and low confidence in the prediction \citep{yuksekgonul2024beyond}. Previously marginalized and excluded groups may have more limited representation in the training data set (e.g. women or other minority groups for whom there were obstacles to obtaining loans). Therefore, this difference in uncertainties and performance is tied to the fairness of the model. In supervised ML settings, some scholars propose a definition of fairness as uncertainty differences between groups \citep{romano2020malice}. Others define uncertainty fairness metrics as susceptibility to noise, missingness, and shifts in data, which is framed as being complementary to other existing fairness metrics \citep{kuzucu2024uncertainty}.

We argue that this problem of uncertainty through a history of unequal inclusion in the data set is especially acute in online problems. Data drifts and lack of robustness among subgroups present a greater challenge in sequential decision-making than in a one-off prediction. This is better modeled in RL, as exploration can adapt to data drifts. While there has been a recent burst of interest in fair RL, much of the existing literature focuses on algorithmic interventions for specific types of bias \textbf{without addressing the holistic interaction of uncertainties across the decision pipeline}. 
Early work in this area examined the theoretical properties of algorithms that adhere to strict definitions of individual fairness \citep{joseph2016, jabbari2017}.
Later authors developed critical frameworks for addressing selective label bias \citep{lakkaraju2017, dearteaga2018, Liu2019, damour2020, kilbertus2020, rateike24, de2025time}, as well as the ``Rashomon Effect'' in risk assessments \citep{coston2021}. 
However, these works primarily treated selective labels as an estimation challenge. Our framework extends this by situating selective feedback within a broader taxonomy, illustrating how it compounds with local prediction uncertainties to structurally exclude marginalized groups.
We build especially on the work of \citet{Wen2021}, who propose methods for fair learning in Markov decision processes (MDPs). However, their formalization fails to account for counterfactual rewards. 

A recent survey of fair RL methods \citep{reuel2024fairness} highlights a gap in existing literature as ``fairness over time'': fairness is often treated as a final objective to be fulfilled or maximised at final time step $t$. Our work departs from this by accounting for fairness in each time step. As discussed in \S \ref{sec:legal}, this distinction is particularly relevant in contexts where temporarily unfair decisions may conflict with non-discrimination requirements, even if long-term outcomes appear fair. We contribute a conceptual framing that highlights how uncertainty asymmetry can shape fairness outcomes over time, and how actively engaging with uncertain regions of the decision space can alter these dynamics.
 
Recent contributions have emphasized causal approaches, acknowledging that structural dependencies between variables have important implications for fairness in RL \citep{Zhang_Bareinboim_2018, nabi2019, Creager2020, Huang2021, wang2025}. We argue that this is especially important in the selective label setting, where optimal policies are defined with respect not just to observed data but to all potential outcomes, as formalized in \S \ref{sec:theory}. 
\citet{puranik2022} and \citet{Hu_Zhang_2022} develop sophisticated optimization policies and causal interventions to achieve long-term fairness targets. Similarly, \citet{yin2023} formalize long-term fairness as an RL problem that adapts to unknown population dynamics. While these approaches are effective at steering a system toward a desired equilibrium, they often require a high degree of control over environmental dynamics or explicit fairness constraints. 
Our proposal is simpler. By accounting for the unequal distribution of uncertainty, specifically the representation bias that inflates the risk profile of under-represented groups, we can improve fairness outcomes without the need for explicit, and potentially legally precarious, affirmative action constraints. Moreover, optimizing over counterfactual outcomes (i.e., unrealized gains and losses) can improve performance as well as fairness. This has been previously reported in supervised learning \citep{zhou2024counterfactual}. Our work suggests that it can hold true in RL settings as well.

\section{Conclusion}\label{sec:concl}

Quantifying and managing uncertainty is a key challenge in many decision-making contexts, especially in reinforcement learning where optimal policies must strike a balance between exploration and exploitation. When political, economic, and institutional policies systematically inflate uncertainty measures for marginalized groups, this technical challenge takes on a newfound social significance. We have argued that many standard approaches to online learning fail to properly account for the biases that can accumulate under feedback uncertainty. This is especially important given the prevalence of sequential decision-making in high-risk settings such as hiring, education, welfare allocation, and finance, in which decisions determine whose outcomes are observed. 

These concerns are not only normative but increasingly emerging in regulatory and legal debates around disparate impact and accountability. However, there are limited established practices on evaluating fairness in sequential systems with missing counterfactuals. Our taxonomy and accompanying experiments suggest that uncertainty-aware algorithmic design can serve as one component of broader socio-technical interventions. The taxonomy provides a structured lens for diagnosing where and how uneven uncertainty enters a decision pipeline, while the experiments illustrate how accounting for that uncertainty can anticipate and mitigate harm from selective feedback.
Importantly, our approach is not proposed as substitutes for other interventions, e.g. institutional reform, contestability, and policies. Rather, we aim to equip practitioners and researchers with a shared vocabulary and a set of tools for evaluating sequential decision pipelines for uncertainty-driven harms. Our simulations illustrate that it is possible to design and implement algorithms that incorporate differential uncertainty across sensitive attributes in a principled way, simultaneously improving outcomes for decision-makers and historically disadvantaged groups. Our approach has potential applications across a wide range of systems in which sequential high-stakes decisions are being made. The taxonomy and uncertainty framing may help practitioners and policymakers reason about potential sources of disparity in sequential decision pipelines.

There are several possible directions for future work in this area. First, the legal status of RL for high risk sequential settings 
such as credit risk assessment and insurance pricing remains unclear. This could be addressed through technically informed policies that acknowledge and attempt to mitigate the various modes of uncertainty that are inherent to probabilistic reasoning. Given the prevalence of and potential for sequential decision-making in high-risk settings, any %
governance-related interventions should be made with care and crafted through a transparent process that brings in stakeholders from multiple sectors.\footnote{These observations should be interpreted as considerations for governance discussions rather than as claims that the proposed methods satisfy legal or regulatory requirements.}

On a technical level, several important open questions remain.
For instance, ideal methods for quantifying and trading off uncertainty measures is an active area of research with direct implications on this study. Understanding the pre-conditions for active exploration within an uncertainty boundary could improve the fairness of RL algorithms. Extending these approaches to the continuous setting is a nontrivial challenge. For example, an insurance offer that was rejected may have been accepted by the customer at a lower price point. 
Greater integration with causal approaches, e.g. using graphical discovery procedures and/or estimating causal effects via adjustment formulae, could also pose a promising direction for further research.
Future work will be required to empirically evaluate how uncertainty-aware auditing and design practices perform in real-world deployments. Our work provides a step in this direction by identifying and formalizing how uncertainty can act as a structural driver of disparity in sequential decision systems.

More broadly, this work contributes to the FAccT community’s effort to move beyond static notions of fairness toward frameworks that account for the dynamics of real-world decision systems. By introducing a taxonomy of uncertainty and illustrating how unequal uncertainty can accumulate under selective feedback, we provide a conceptual lens for diagnosing and reasoning about fairness risks in sequential decision pipelines. We hope this perspective helps researchers, practitioners, and policymakers identify new questions, auditing practices, and design choices for governing uncertainty-driven harms in socio-technical systems.

\begin{acks}
DW was supported by EPSRC grant number UKRI918.
\end{acks}

\section*{Generative AI Usage Statement}

The generative AI tool ChatGPT was used to help with fine tuning the visual elements of the Matplotlib~\citep{Hunter:2007} code which is used for the plots. 

\bibliographystyle{ACM-Reference-Format}
\bibliography{references}

\appendix

\section{Synthetic data}
\label{app:data}

The underlying phenomenon takes the following form:
\begin{align*}
Y &= f(X)\;+\;\varepsilon\\
X^{\mathrm{obs}} &= g(X)\\
Y^{\mathrm{obs}} &= h(Y),
\end{align*}
where $X$ denotes the latent covariates, and $X^{\mathrm{obs}}, Y^{\mathrm{obs}}$ are their potentially biased observations. The main variables in this framework are:
\begin{itemize}
  \item $A\in\{0,1\}$ – binary sensitive attribute;  
  \item $R\in\mathbb R_{\ge 0}$ – \emph{resources} (continuous, e.g.\ salary);  
  \item $Q\in\{0,\dots,K\}$ – contextual/discrete feature (e.g.\ city zone);  
  \item $S\in\mathbb R$ – latent score driving the binary target;  
  \item $Y=\mathbf 1\!\{S>\Pi_S\}$ – ground-truth label, with threshold $\Pi_S$.  
\end{itemize}

In our interpretation of the variables in the context of granting loans, the continuous variable $R$ is interpreted as the salary, the discrete $Q$ is the city zone, and $A$ is the sensitive attribute. The following equations define the relationships between the variables and the knobs that are tweaked to get different magnitudes of different types of bias.
\begin{align*}
A          &=  B_A,               & B_A &\sim\mathrm{Ber}(p_A);                                                          &&\text{(4a)}\\
R          &= -\beta_R^{(h)}\,A + N_R,             & N_R &\sim\mathrm{Gamma}(k_R,\theta_R);                                   &&\text{(4b)}\\
Q          &=  B_Q,               & B_Q\mid(R,A) &\sim\mathrm{Bin}\!\bigl(K,\;p_Q(R,A)\bigr), \nonumber\\[-2pt]
           &&&\displaystyle p_Q(R,A)=\sigma\!\bigl(-(\alpha_{RQ}R-\beta_Q^{(h)}A)\bigr);                              &&\text{(4c)}\\
S          &=  \alpha_R R - \alpha_Q Q - \beta_Y^{(h)} A + N_S, & N_S &\sim\mathcal N(0,\sigma_S^2);                     &&\text{(4d)}\\
Y          &=  \mathbf 1\!\{S>\Pi_S\}.                                                                                        &&&&\text{(4e)}
\end{align*}
Here, historical bias is introduced via the coefficients $\beta_R^{(h)}$, $\beta_Q^{(h)}$, and $\beta_Y^{(h)}$, which reduce resources, contextual quality, or score based on sensitive group membership ($A=1$). To simulate measurement errors or biased proxies in observed data, additional noise and group-dependent offsets are applied as follows:
\begin{align*}
\tilde R   &= R - \beta_R^{(m)} A + N_{\tilde R}, & N_{\tilde R} &\sim \mathcal N(0,\sigma_{\tilde R}^2);        &&\text{(5a)}\\
\tilde S   &= S - \beta_Y^{(m)} A + N_{\tilde S}, & N_{\tilde S} &\sim \mathcal N(0,\sigma_{\tilde S}^2);        &&\text{(5b)}\\
\tilde Y   &= \mathbf 1\!\{\tilde S>\Pi_S\}.                                                                                 &&&&\text{(5c)}
\end{align*}
The coefficients $\beta_R^{(m)}$ and $\beta_Y^{(m)}$ represent additional biases in recorded variables, such as under reporting of resources or distortion in outcome recording for specific groups.

\xhdr{Overview of bias parameters}
\begin{itemize}
  \item \emph{Historical bias} enters through the shift parameters $\beta_R^{(h)},\beta_Q^{(h)},\beta_Y^{(h)}$ in Eqs.\,(5b)–(5d).
  \item \emph{Measurement bias} is governed by $\beta_R^{(m)},\beta_Y^{(m)}$ in the proxy equations (6a)–(6c).
  \item \emph{Representation bias} can be induced by altering the sampling fraction of $A$ or by manipulating $p_Q(R,A)$, while
        \emph{omitted-variable bias} is simulated by intentionally removing $R$ (or another relevant feature) from the training set.
\end{itemize}

Here $\sigma(\cdot)$ denotes the logistic sigmoid, $\mathrm{Ber}$ and $\mathrm{Bin}$ are Bernoulli and Binomial distributions, and all noise terms are mutually independent.  Setting the corresponding $\beta$’s to zero recovers an unbiased data-generating process; increasing their magnitude amplifies the chosen type(s) of bias.  This compact system is sufficiently flexible to combine several biases at once while remaining analytically transparent.

We refer readers to the original paper covering the synthetic data generation pipeline we leverage \cite{baumann2023bias} and accompanying GitHub repository\footnote{\url{https://github.com/rcrupiISP/BiasOnDemand/tree/main}.} for further details.

\section{Evaluation metrics}
\label{app:evaluation}

All fairness metrics are computed with respect to $Y_\text{real}$, the unbiased ground-truth repayment outcome, not the potentially biased observed label $Y$. For a sensitive attribute $A \in \{0,1\}$ and predicted decision $\hat{D}$, we define:
\begin{itemize}
  \item Selection rate: $\mathrm{SR}(a) = P(\hat{D}=1 \mid A=a)$; plotted as the difference $\mathrm{SR}(0) - \mathrm{SR}(1)$.
  \item False positive rate (FPR): $P(\hat{D}=1 \mid Y_\text{real}=0, A=a)$; plotted as $\mathrm{FPR}(0) - \mathrm{FPR}(1)$.
  \item False negative rate (FNR): $P(\hat{D}=0 \mid Y_\text{real}=1, A=a)$; plotted as $\mathrm{FNR}(0) - \mathrm{FNR}(1)$.
  \item Utility: normalized cumulative loan profit per quarter.
\end{itemize}
Differences closer to zero indicate better group parity; negative values indicate the minority group ($A=0$) is disadvantaged relative to the majority.

\section{Further Experiments}
\label{app:experiments}

This appendix presents three complementary views of method performance across the full configuration sweep.
Section~\ref{app:heatmaps} provides method-ranking heatmaps that compare all five methods at the final decision round under each bias condition individually.
Section~\ref{app:temporal} traces fairness metrics and profit over time to reveal how methods converge, overshoot, or diverge across quarters.
Section~\ref{app:violin} aggregates over all bias conditions simultaneously via violin plots, exposing the distributional spread,and especially the worst-case tails,of each method's outcomes.
Together, these analyses reinforce that no single method dominates: the appropriate uncertainty-aware strategy depends on the structure of the bias present. And that there are many scenarios where it could be worth it having an uncertainty aware method.

Table~\ref{tab:sweep-params} lists every parameter used in the full configuration sweep (\texttt{synthetic-run-1.csv}).
Swept parameters are those varied independently; all combinations are crossed, giving a total of
$3^6 \times 3 \times 3 \times 5 = 10{,}935$ configurations (before fixed-parameter dimensions).
Fixed parameters are held constant across all runs.
Bias parameters use the notation of Appendix~\ref{app:data}; method hyperparameters control the sequential decision procedure.

\begin{table}[ht]
\centering
\caption{Parameter sweep for the plots.  Swept parameters are varied independently and fully crossed; fixed parameters are constant across all runs.}
\label{tab:sweep-params}
\small
\begin{tabular}{llll}
\toprule
\textbf{Parameter} & \textbf{Symbol / key} & \textbf{Values} & \textbf{Description} \\
\midrule
\multicolumn{4}{l}{\textit{Swept — bias parameters}} \\
\midrule
Label historical bias      & $\beta_Y^{(h)}$ / \texttt{l\_y}   & 0, 2, 4          & Shifts latent score downward for $A{=}1$ \\
Label measurement bias     & $\beta_Y^{(m)}$ / \texttt{l\_m\_y} & 0, 2, 4          & Distorts observed label for $A{=}1$ \\
Resource historical bias   & $\beta_R^{(h)}$ / \texttt{l\_h\_r} & 0, 1, 4          & Reduces observed resource $R$ for $A{=}1$ \\
$Q$ historical bias        & $\beta_Q^{(h)}$ / \texttt{l\_h\_q} & 0, 1, 4          & Reduces contextual feature $Q$ for $A{=}1$ \\
Measurement / proxy bias   & $\beta_R^{(m)}$ / \texttt{l\_m}    & 0, 1, 2          & Replaces $R$ with biased proxy $\tilde{R}$ when $>0$ \\
Interaction proxy bias     & $\beta_{Yb}$    / \texttt{l\_y\_b} & 0, 2, 4          & Interaction term between proxy and group membership \\
\midrule
\multicolumn{4}{l}{\textit{Swept — method hyperparameters}} \\
\midrule
Certain-applicant fraction & \texttt{proportion\_certain} & 0.6, 0.7, 0.8 & Share of loan budget allocated to certain (non-explored) applicants \\
Exploration fraction       & $\delta$ / \texttt{delta}    & 0.02, 0.05, 0.1 & Fraction of each quarter reserved for uncertainty-driven exploration \\
Random seed                & \texttt{random\_seed}         & 0, 20, 40, 60, 80 & Controls data generation and model initialisation \\
\midrule
\multicolumn{4}{l}{\textit{Fixed — data generation}} \\
\midrule
Dataset size               & \texttt{dim}              & 20\,000   & Total number of simulated applicants \\
Number of decision rounds  & \texttt{num\_partitions}  & 8         & Quarters into which the data is split sequentially \\
Discrete feature levels    & \texttt{l\_q}             & 2         & Number of levels for $Q$ \\
Score noise std.\ dev.\    & \texttt{sy}               & 2         & $\sigma_S$ in the latent score equation \\
\midrule
\multicolumn{4}{l}{\textit{Fixed — decision procedure}} \\
\midrule
Rejection threshold        & \texttt{rejection\_threshold} & 0.1  & Minimum predicted probability to grant a loan \\
Budget proportion          & \texttt{budget\_prop}         & 0.8  & Fraction of ground-truth positive loans used as budget \\
Gain percentage            & \texttt{gain\_percentage}     & 0.4  & Profit margin on each repaid loan \\
\midrule
\multicolumn{4}{l}{\textit{Fixed — model training}} \\
\midrule
Training epochs            & \texttt{n\_epochs} & 40   & Epochs per quarter for the logistic regression update \\
Learning rate              & \texttt{lr}        & 0.05 & Gradient-descent step size \\
\bottomrule
\end{tabular}
\end{table}

\subsection{Method-Ranking Heatmaps}
\label{app:heatmaps}

\begin{figure}[ht]
  \centering
  \begin{subfigure}[t]{0.48\textwidth}
    \centering
    \includegraphics[width=\textwidth]{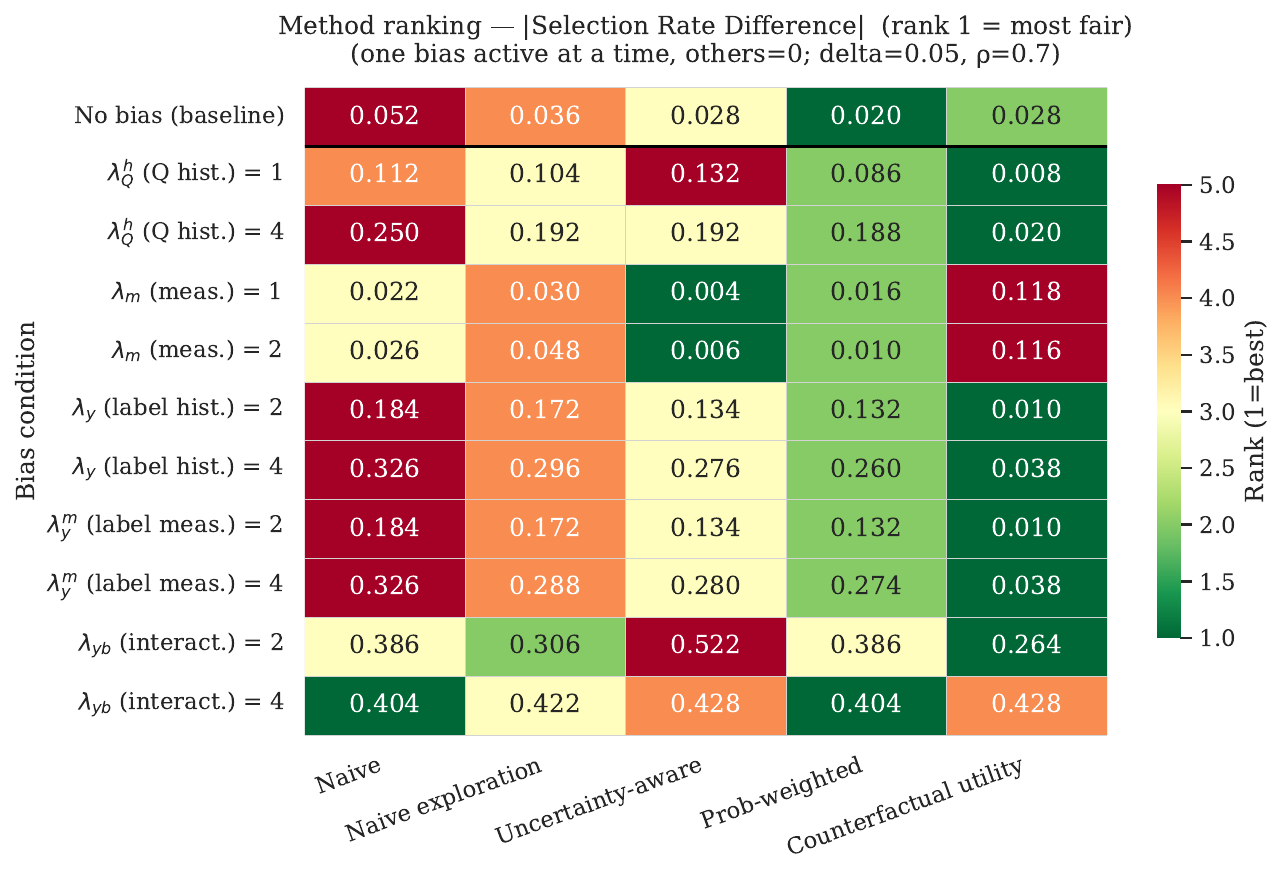}
    \caption{$|\Delta\text{SR}|$ (selection-rate difference)}
    \label{fig:heatmap-sr}
  \end{subfigure}
  \hfill
  \begin{subfigure}[t]{0.48\textwidth}
    \centering
    \includegraphics[width=\textwidth]{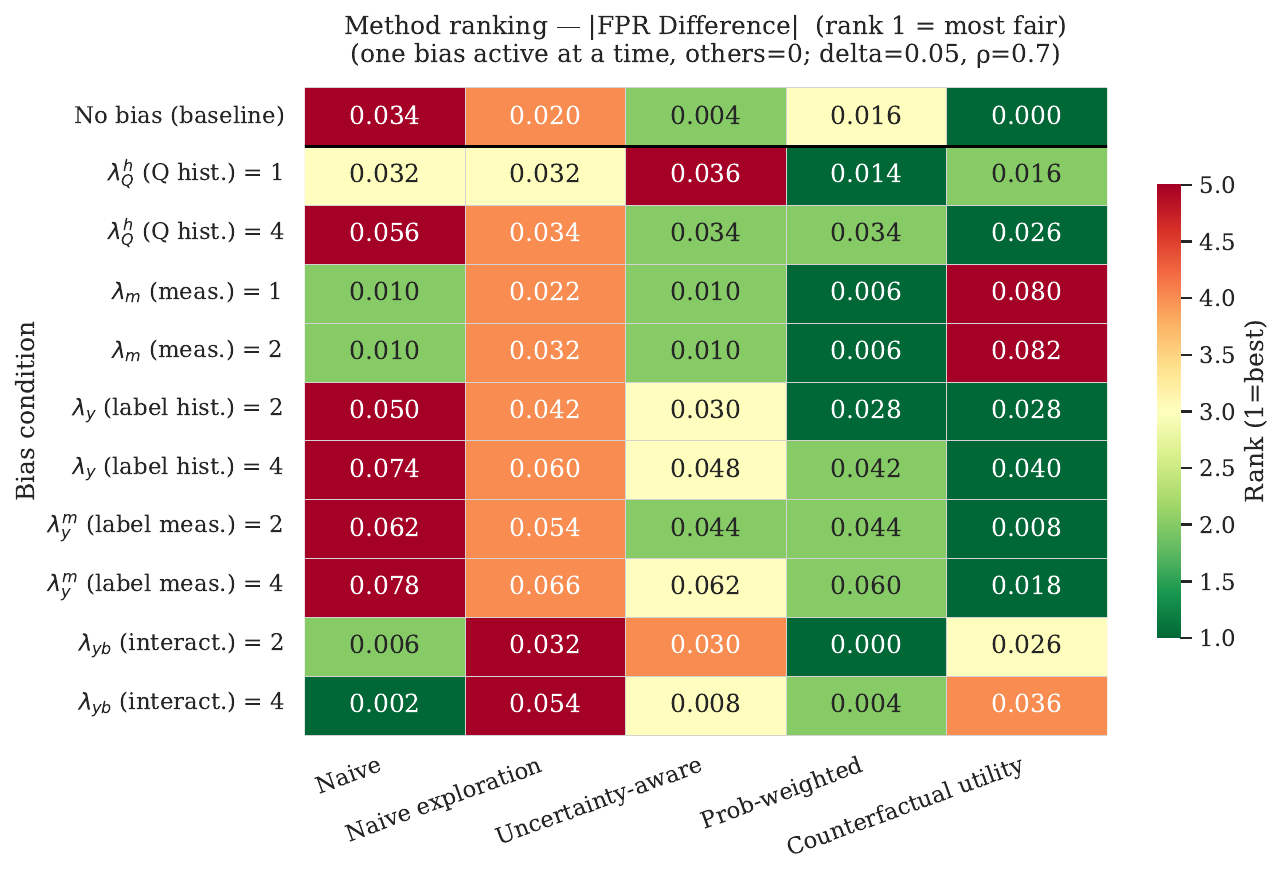}
    \caption{$|\Delta\text{FPR}|$ (false-positive-rate difference)}
    \label{fig:heatmap-fpr}
  \end{subfigure}

  \vspace{0.5em}

  \begin{subfigure}[t]{0.48\textwidth}
    \centering
    \includegraphics[width=\textwidth]{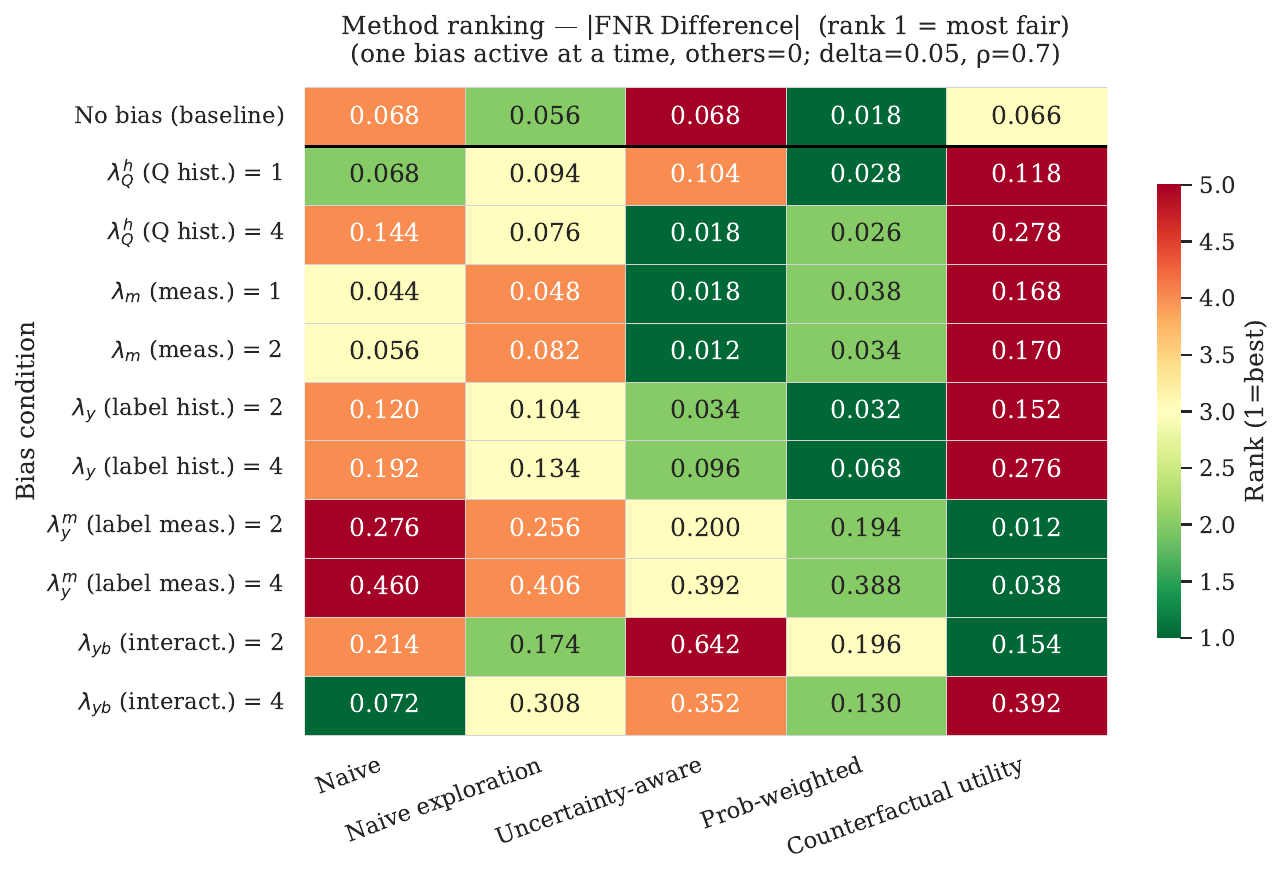}
    \caption{$|\Delta\text{FNR}|$ (false-negative-rate difference)}
    \label{fig:heatmap-fnr}
  \end{subfigure}
  \hfill
  \begin{subfigure}[t]{0.48\textwidth}
    \centering
    \includegraphics[width=\textwidth]{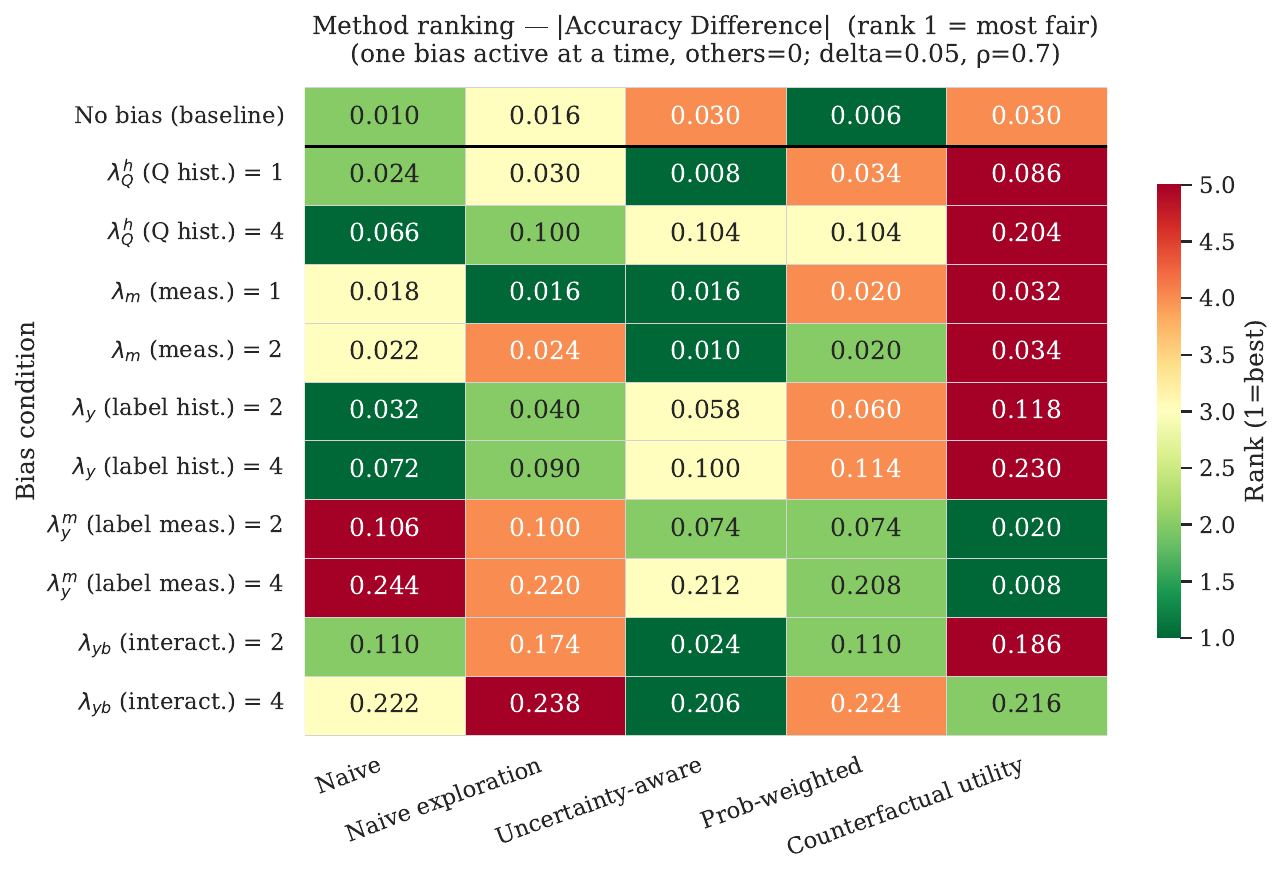}
    \caption{$|\Delta\text{Acc}|$ (accuracy difference)}
    \label{fig:heatmap-acc}
  \end{subfigure}

  \caption{%
    Method rankings by four fairness metrics (lower disparity = better; rank~1 = best).
    No single method dominates all bias conditions.
    Counterfactual utility ranks well under label bias ($\beta_Y^{(h)}$, $\beta_Y^{(m)}$) but worst under proxy measurement bias ($\beta_R^{(m)}$).
    Under interaction proxy bias ($\beta_{Yb}$), all methods show substantial disparity.
    Panel~(c) is the most condition-dependent, with method rankings inverting across bias types.
  }
  \label{fig:heatmap-fairness}
\end{figure}

\begin{figure}[ht]
  \centering
  \includegraphics[width=0.55\textwidth]{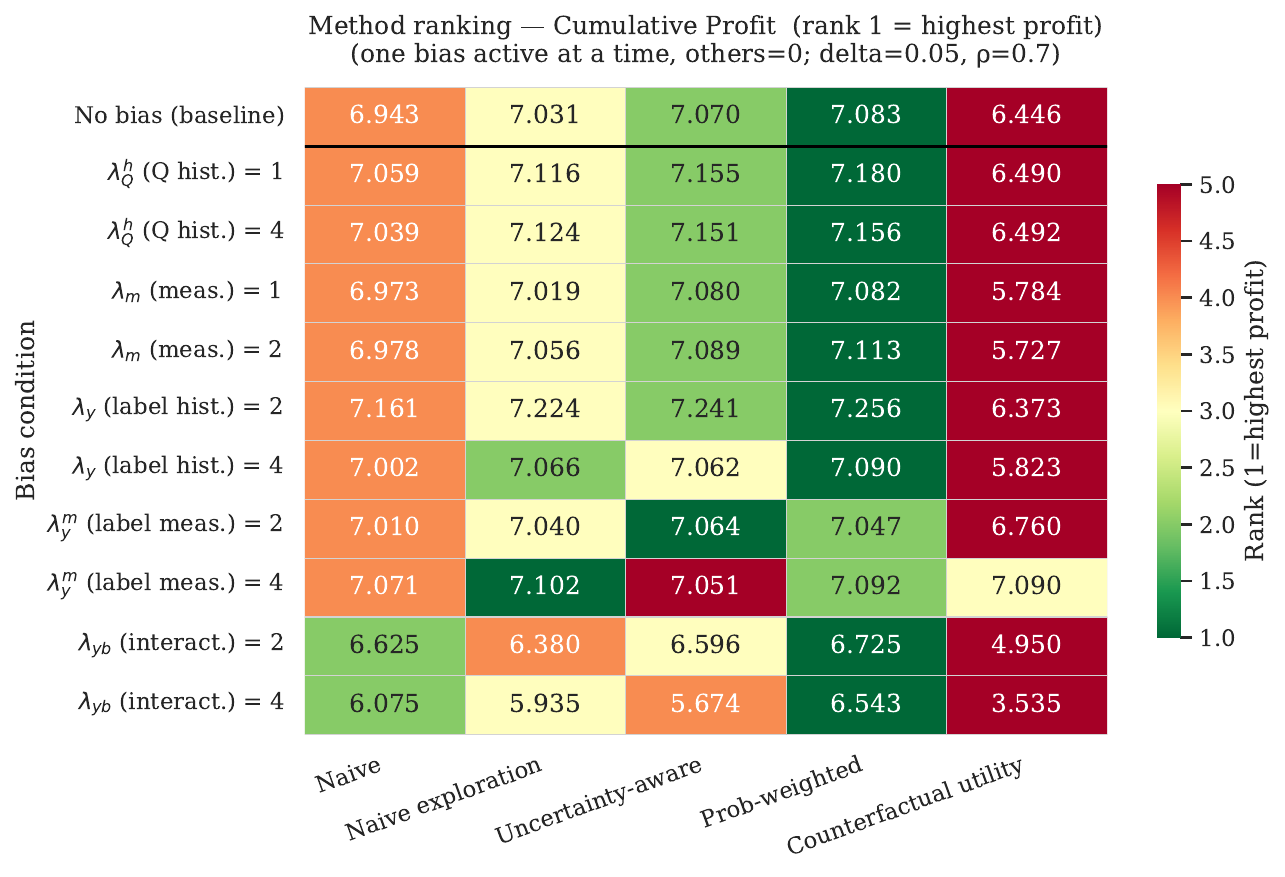}
  \caption{%
    Method ranking by cumulative profit (higher = better; rank~1 = most profitable).
    Prob-weighted exploration ranks first in nearly every condition.
    Counterfactual utility is consistently the least profitable, reflecting its fairness--profit trade-off.
    The naive baseline is not the most profitable strategy either, as its lack of minority-group data limits long-run model calibration.
  }
  \label{fig:heatmap-profit}
\end{figure}

The heatmaps in Figures~\ref{fig:heatmap-fairness} and~\ref{fig:heatmap-profit} summarise how the five methods rank relative to each other across every bias condition tested in a sweep of the bias configurations.
Each row corresponds to a single bias parameter varied at one level while all other bias parameters are fixed at zero, plus a no-bias baseline row (top, separated by a horizontal rule) in which all $\beta$ coefficients are zero.
The five columns correspond to the five methods.

Cell colour encodes rank: green = rank~1 (best), red = rank~5 (worst), with ties sharing the lowest rank.
Cell annotations show the underlying metric value averaged over random seeds, so that ties or near-ties in rank are still visually distinguishable.
All results are taken at the final decision round (Q10) with fixed hyperparameters $\delta = 0.05$ and \texttt{proportion\_certain} $= 0.7$.

Bias parameters follow the notation of Appendix~\ref{app:data}:
$\beta_Y^{(h)}$ (historical bias on label), $\beta_Y^{(m)}$ (measurement bias on label), $\beta_Q^{(h)}$ (historical bias on feature $Q$), $\beta_R^{(m)}$ (measurement bias inducing proxy $\tilde{R}$), and $\beta_{Yb}$ (interaction proxy bias).

A method that is consistently green across all rows is robust to diverse bias types; a method that is green for some rows but red for others is sensitive to the specific bias mechanism.
The baseline row provides a sanity check: in the absence of bias all methods should perform similarly, so ranks should be close to a tie.

\subsection{Temporal Fairness Recovery}
\label{app:temporal}

\begin{figure}[ht]
  \centering
  \includegraphics[width=\textwidth]{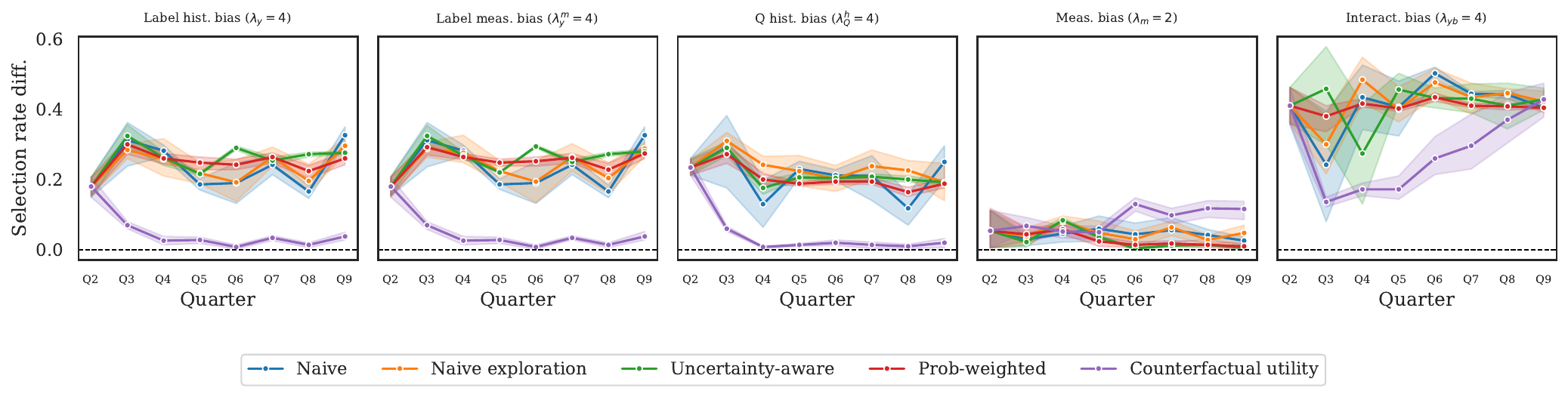}
  \caption{%
    Selection-rate difference ($\Delta\text{SR}$) over decision rounds under each bias type at maximum severity.
    Counterfactual utility converges toward parity under label bias but diverges under proxy measurement bias ($\beta_R^{(m)}$), where its utility estimate inherits the proxy corruption.
    Under interaction proxy bias ($\beta_{Yb}$), all methods show persistent disparity.
  }
  \label{fig:temporal-sr}
\end{figure}

\begin{figure}[ht]
  \centering
  \includegraphics[width=\textwidth]{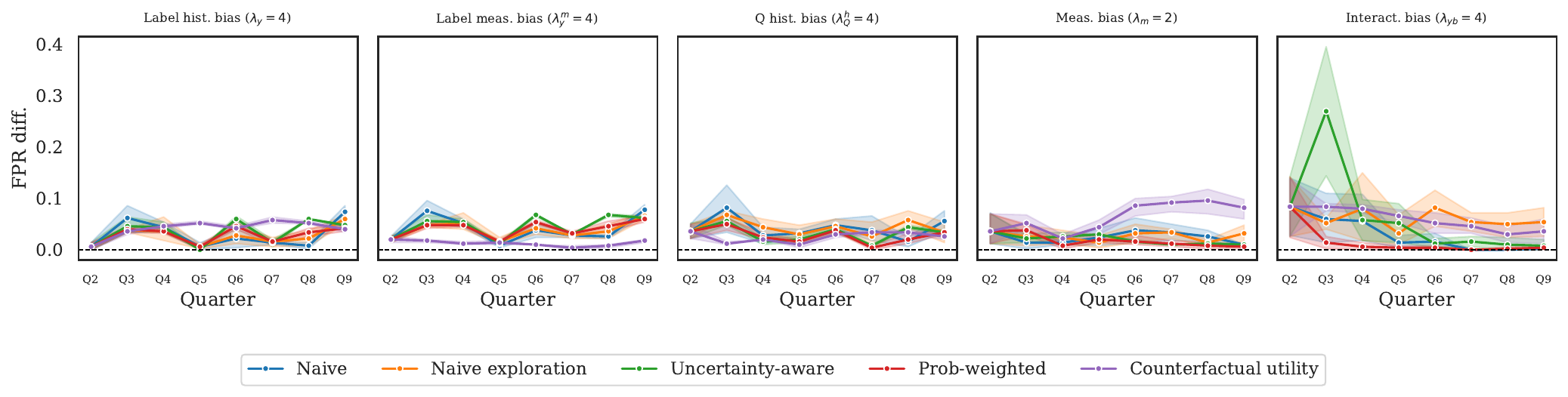}
  \caption{%
    False-positive-rate difference ($\Delta\text{FPR}$) over decision rounds.
    Under label bias, counterfactual utility reduces $\Delta\text{FPR}$; under proxy measurement bias, prob-weighted exploration achieves the smallest gap.
    Which method controls FPR best depends on the bias mechanism.
  }
  \label{fig:temporal-fpr}
\end{figure}

\begin{figure}[ht]
  \centering
  \includegraphics[width=\textwidth]{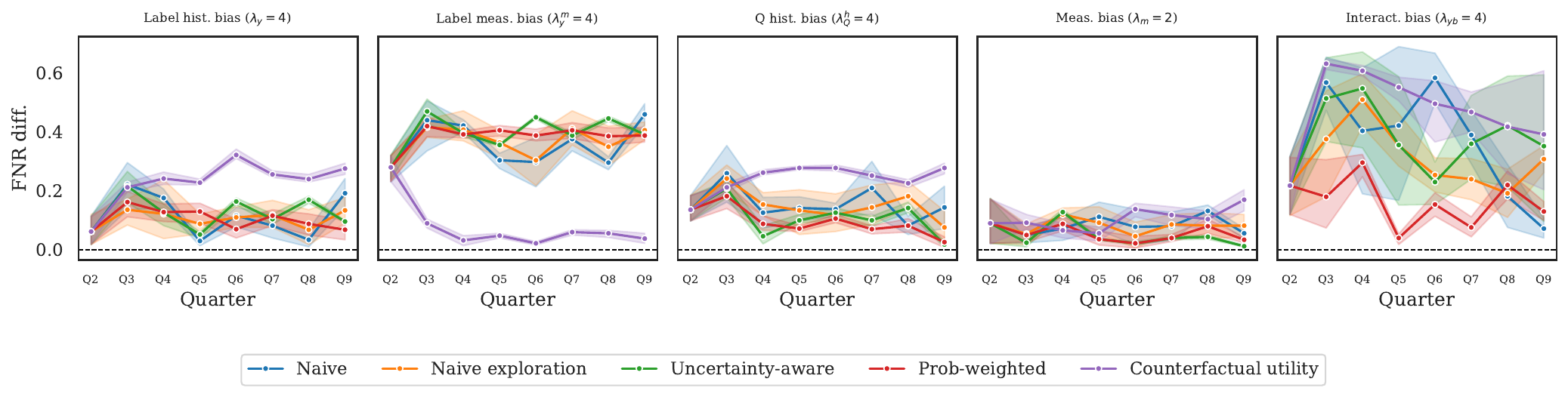}
  \caption{%
    False-negative-rate difference ($\Delta\text{FNR}$) over decision rounds.
    Trajectories are more unstable than for $\Delta\text{SR}$ or $\Delta\text{FPR}$: methods that increase minority selection can initially reduce $\Delta\text{FNR}$ but overshoot in later quarters.
    Under interaction proxy bias, trajectories diverge rather than converge, illustrating the limits of uncertainty treatment under structural proxy corruption.
  }
  \label{fig:temporal-fnr}
\end{figure}

\begin{figure}[ht]
  \centering
  \includegraphics[width=\textwidth]{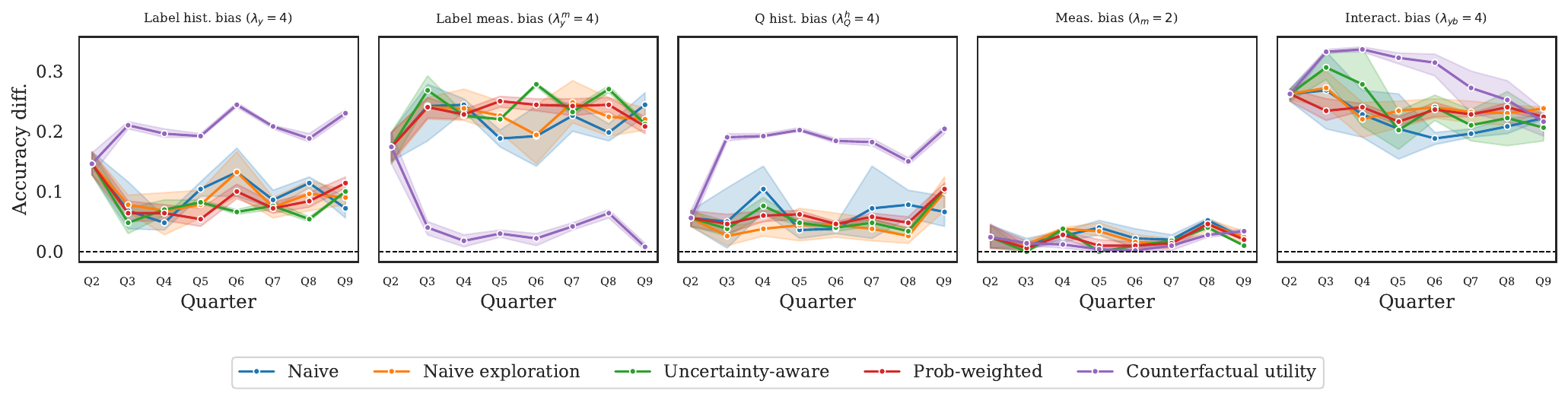}
  \caption{%
    Accuracy difference ($\Delta\text{Acc}$) over decision rounds.
    Under label historical bias, the naive baseline maintains the smallest gap; under label measurement bias, counterfactual utility converges to near-zero $\Delta\text{Acc}$.
    These opposing dynamics reflect that $\Delta\text{Acc}$ responds to changes in both decision policy and label quality.
  }
  \label{fig:temporal-acc}
\end{figure}

\begin{figure}[ht]
  \centering
  \includegraphics[width=\textwidth]{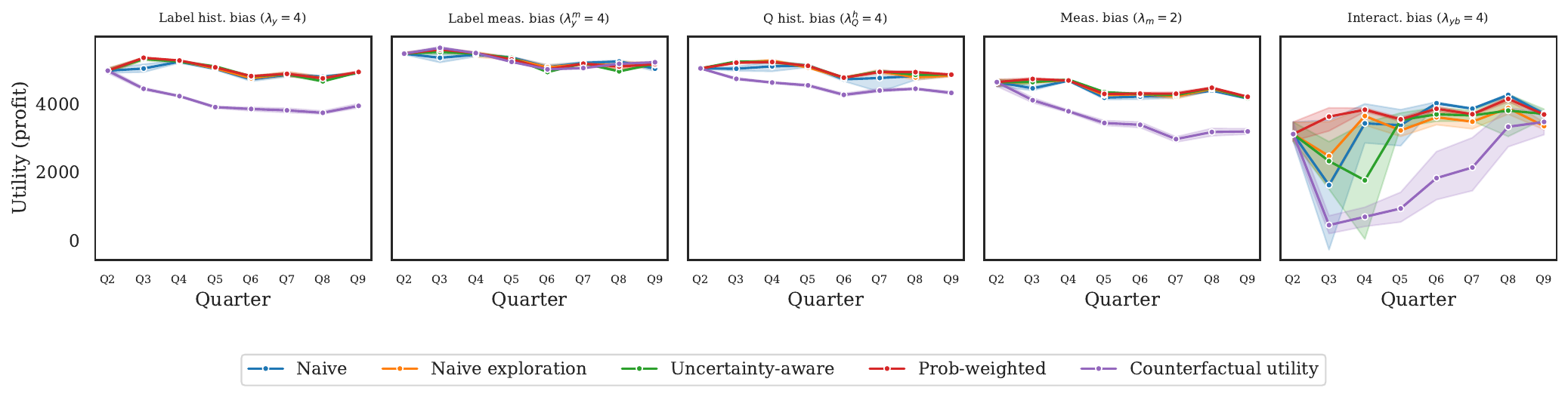}
  \caption{%
    Per-quarter utility (profit) over decision rounds.
    Prob-weighted exploration maintains the highest or near-highest utility across all bias types.
    Counterfactual utility yields the lowest per-quarter profit, reflecting its fairness--profit trade-off.
    Trajectories stabilise after the first few quarters, suggesting that long-run profit differences are largely determined by how uncertainty is handled during initial data collection.
  }
  \label{fig:temporal-utility}
\end{figure}

Figures~\ref{fig:temporal-sr}--\ref{fig:temporal-utility} show how fairness metrics and profit evolve quarter-by-quarter for each of the five bias types at maximum severity (all other bias parameters fixed at zero).
Each panel corresponds to one bias type; lines represent the five methods; the dashed horizontal line at zero marks perfect group parity.
Hyperparameters are fixed at $\delta=0.05$ and \texttt{proportion\_certain}$=0.7$.

These plots illustrate trajectories rather than final-state comparisons.
Accounting for uncertainty,and choosing the form of uncertainty treatment,changes the dynamic path a system takes, not only its final-quarter snapshot.
A method may reach parity faster, overshoot it, or oscillate, depending on the structure of the bias it faces.
We make no claim that any single method is universally preferable; rather, the figures show that the nature of the uncertainty present in the data shapes which strategy is appropriate.

\subsection{Worst-Case Distribution of Outcomes}
\label{app:violin}

\begin{figure}[ht]
  \centering
  \includegraphics[width=0.75\textwidth]{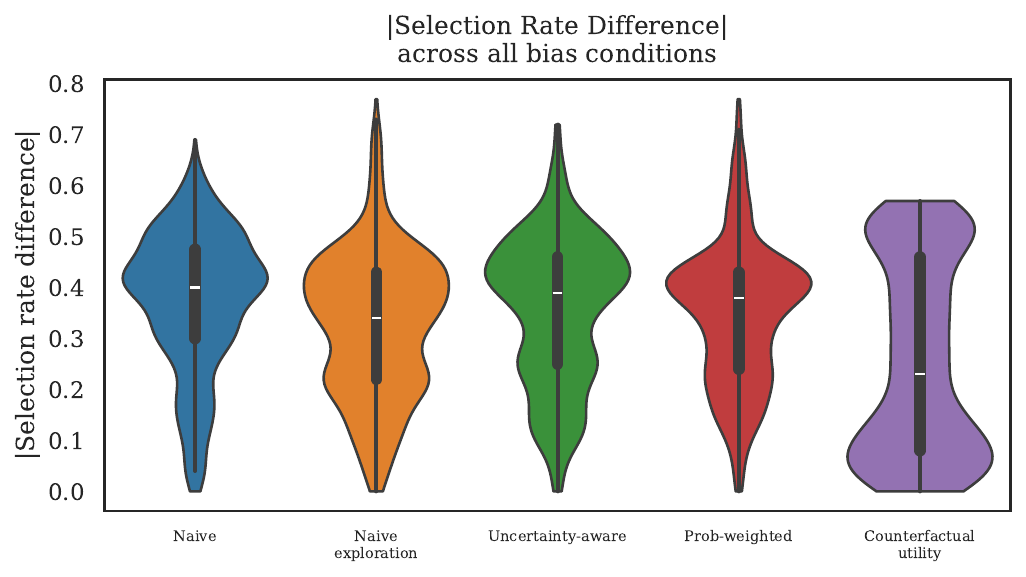}
  \caption{%
    Distribution of $|\Delta\text{SR}|$ at the final quarter across all bias conditions and random seeds.
    Uncertainty-aware and prob-weighted exploration show narrower, lower-positioned violins than the naive baseline.
    Counterfactual utility has a bimodal distribution: concentrated near zero under label bias but with a long upper tail under proxy bias.
  }
  \label{fig:violin-sr}
\end{figure}

\begin{figure}[ht]
  \centering
  \includegraphics[width=0.75\textwidth]{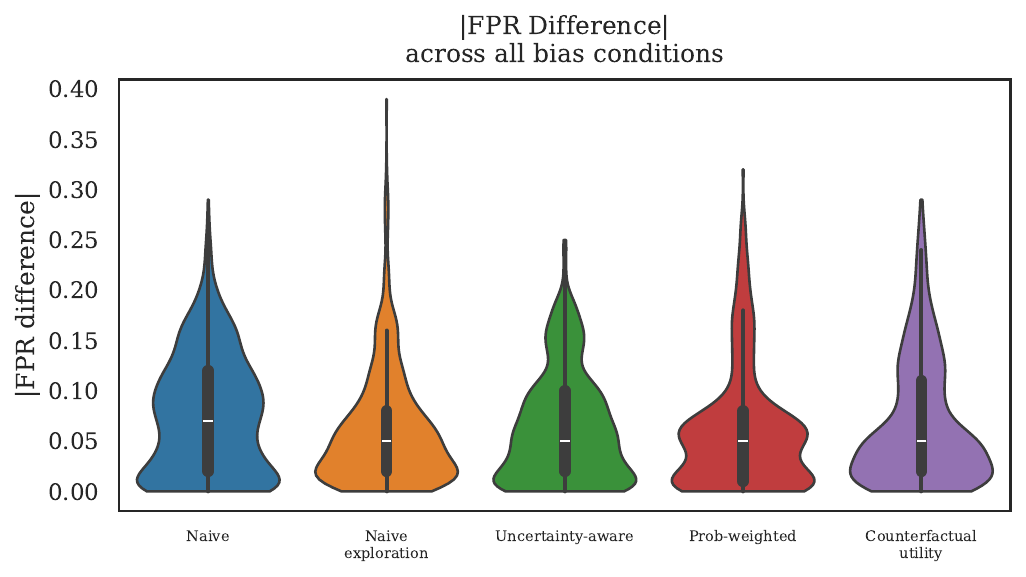}
  \caption{%
    Distribution of $|\Delta\text{FPR}|$ across all bias conditions.
    Prob-weighted exploration produces the most compact violin.
    The naive baseline has a wider distribution, and the tails,not the medians,are where methods diverge most.
  }
  \label{fig:violin-fpr}
\end{figure}

\begin{figure}[ht]
  \centering
  \includegraphics[width=0.75\textwidth]{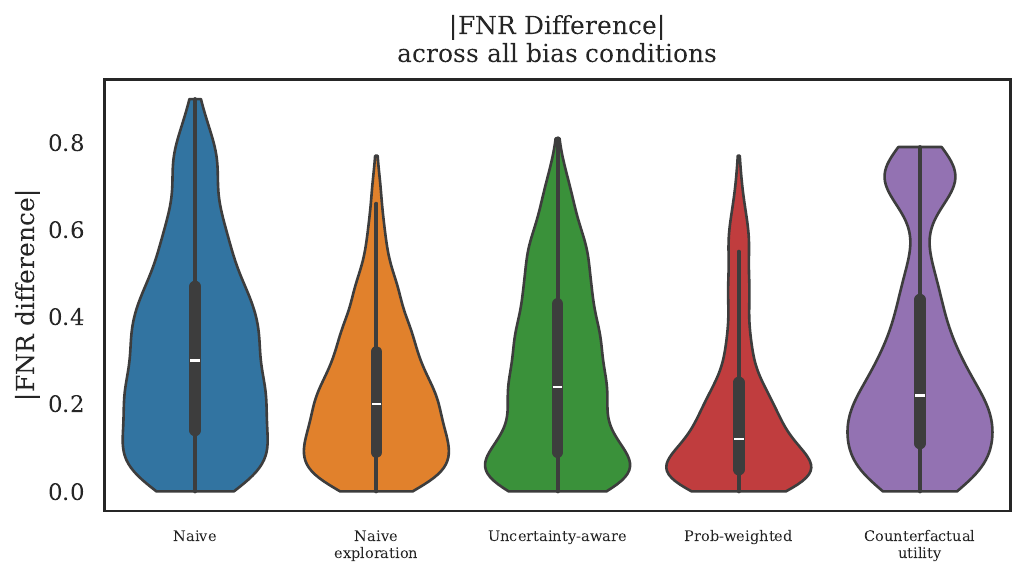}
  \caption{%
    Distribution of $|\Delta\text{FNR}|$ across all bias conditions.
    FNR disparity is the most volatile metric: no method fully controls the upper tail.
    Counterfactual utility has the widest distribution.
    The broad spread across all methods suggests that FNR disparity is harder to control without explicit constraints.
  }
  \label{fig:violin-fnr}
\end{figure}

\begin{figure}[ht]
  \centering
  \includegraphics[width=0.75\textwidth]{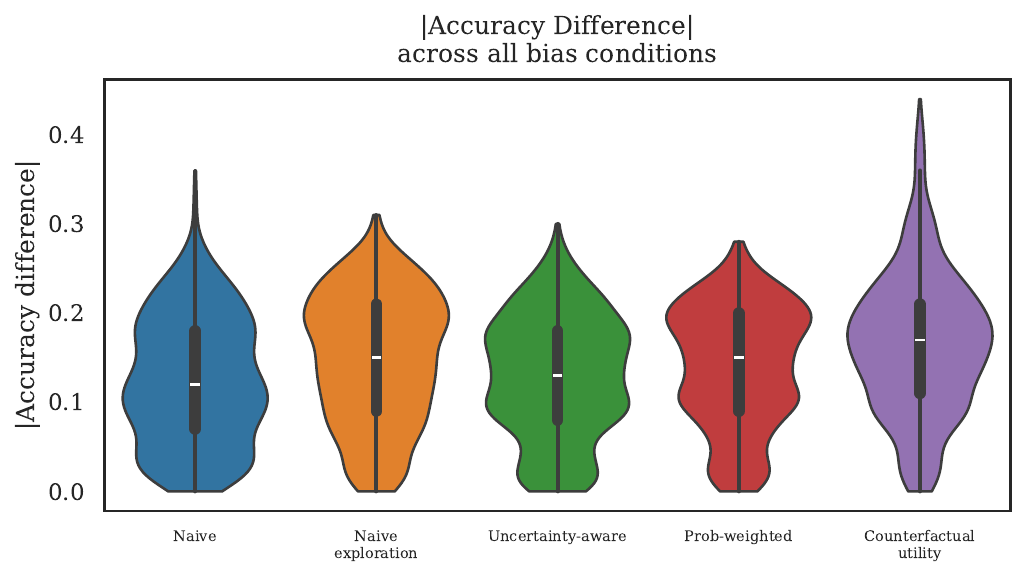}
  \caption{%
    Distribution of $|\Delta\text{Acc}|$ across all bias conditions.
    Distributions are narrower than for FNR disparity, as accuracy aggregates false positives and false negatives that partially cancel.
    Uncertainty-aware and prob-weighted exploration show lower medians than the naive baseline; counterfactual utility shows the widest spread.
  }
  \label{fig:violin-acc}
\end{figure}

\begin{figure}[ht]
  \centering
  \includegraphics[width=0.75\textwidth]{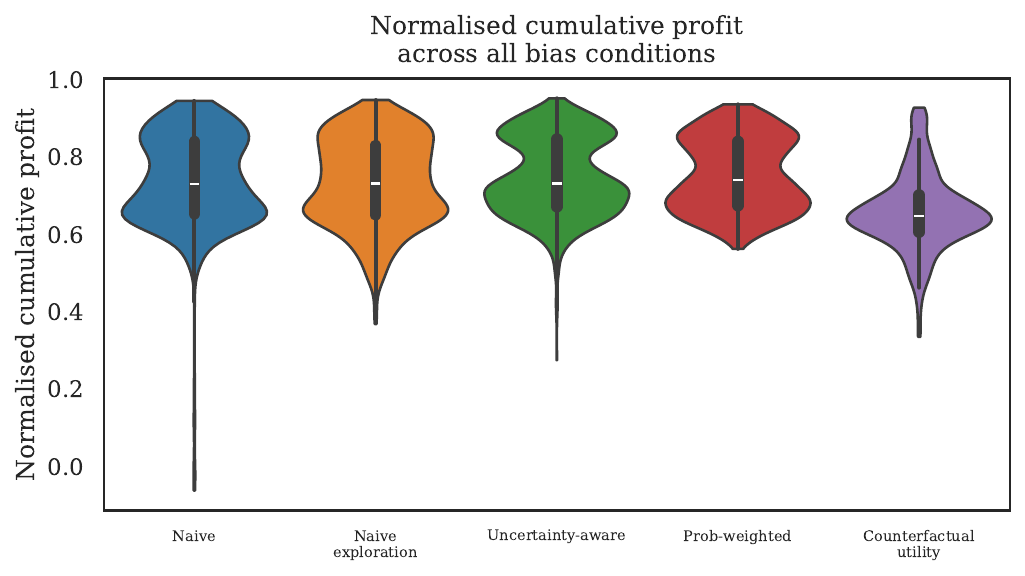}
  \caption{%
    Distribution of normalised cumulative profit across all bias conditions.
    Prob-weighted exploration achieves the highest median with the narrowest distribution.
    Counterfactual utility has the lowest median, reflecting its fairness--profit trade-off.
    The naive baseline falls between, with a wider spread indicating greater sensitivity to the bias configuration.
  }
  \label{fig:violin-profit}
\end{figure}

The violin plots in Figures~\ref{fig:violin-sr}--\ref{fig:violin-profit} show the distribution of each metric at the final decision round across all bias conditions and random seeds simultaneously,no filtering by bias type.
Each violin aggregates over the full combinatorial sweep, providing a worst-case view: a method with a narrow, low-positioned violin is reliably fair across diverse bias environments, while a wide or high-tailed violin indicates fragility.

These plots are not intended to rank methods but to surface the variability of outcomes.
A method that performs well on average may have a long upper tail representing poor outcomes under specific bias configurations.
The central observation is that accounting for uncertainty,without any explicit fairness constraint,can reduce both the mean and the variance of group disparity, but the magnitude of this effect depends on whether the uncertainty in the system is epistemic (reducible through additional data) or structural (arising from proxy corruption or interaction effects).

\end{document}